%% file: ms.tex
\pdfoutput=1

\documentclass[journal]{IEEEtran}

\usepackage{cite}

\ifCLASSINFOpdf
  \usepackage[pdftex]{graphicx}

\else

\fi

\usepackage{amsmath}

\ifCLASSOPTIONcompsoc
  \usepackage[caption=false,font=normalsize,labelfont=sf,textfont=sf]{subfig}
\else
  \usepackage[caption=false,font=footnotesize]{subfig}
\fi

\usepackage{url}

\usepackage{amssymb}
\usepackage{color}
\usepackage{pgfplots}
\usepackage{tikz}
\usepgfplotslibrary{groupplots}
\newlength\figurewidth
\newlength\figureheight

\begin{document}

\title{In-Hand Object Stabilization by\\ Independent Finger Control}

\author{Filipe Veiga,
			Benoni B. Edin,
			and Jan Peters
\thanks{Filipe Veiga and Jan Peters are with the Technische Universit\"{a}t Darmstadt, Darmstadt, Germany, FG Intelligent Autonomous Systems. e-mail:  \{veiga, peters\}@ias.tu-darmstadt.de}%
\thanks{Benoni B. Edin are Ume\r{a} University, Ume\r{a}, Sweden, Department of Integrative Medical Biology. e-mail:  benoni.edin@umu.se}%
\thanks{Jan Peters is also with the Max-Planck-Institut f\"{u}r Intelligente Systeme, T\"{u}bingen, Germany.}
}

\markboth{Submitted to IEEE Transactions on Robotics Journal}%
{Veiga {\textit{et al.}}: In-Hand Object Stabilization by Independent Finger Control}

\maketitle

\input{abstract}

\begin{IEEEkeywords}
in-hand manipulation, modular control, reactive control, tactile feedback, independent finger control, slip prediction.
\end{IEEEkeywords}


\input{Introduction}

\input{PredictingSlip}
\input{ExperimentalEvaluation}
\input{DiscussionConclusion}
\input{Acknowledgements}

\ifCLASSOPTIONcaptionsoff
  \newpage
\fi

\bibliographystyle{IEEEtran}
\bibliography{bibliography}

\begin{IEEEbiography}[{\includegraphics[width=1in,height=1.25in,clip,keepaspectratio]{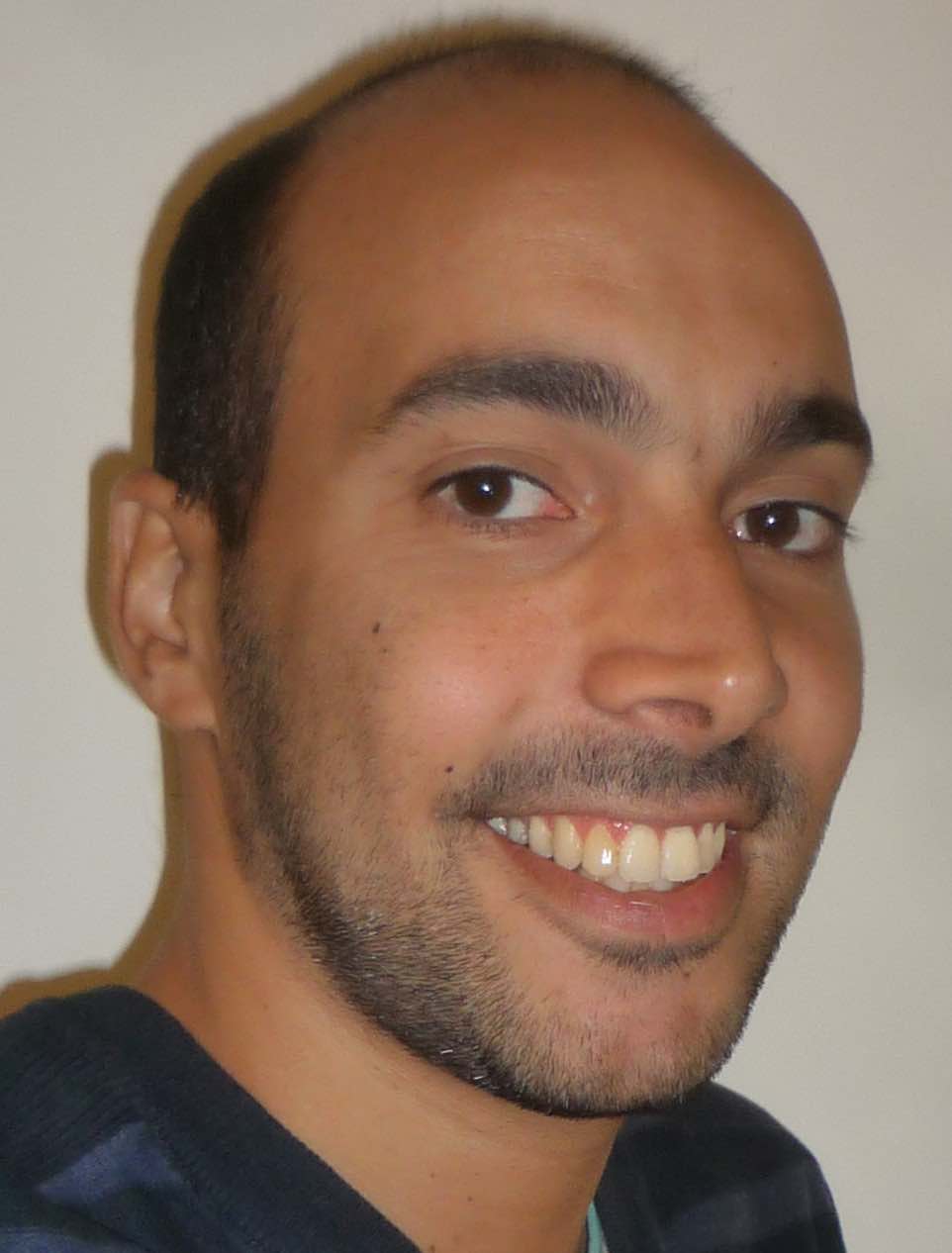}}]{Filipe Veiga}
joined the Intelligent Autonomous System lab of Technische Universitaet Darmstadt on September 1st, 2013 as a PhD student. During his PhD his work on Machine Learning
for Robot Grasping, Manipulation and tactile Exploration under the supervision of Jan Peters focuses on improving the dexterous capability of robots and is part of the TACMAN Project.
Before that, he got his Masters Degree in Electrical and Computer Engineering at Instituto Superior T\'ecnico in Lisbon, Portugal, where his master thesis focused on Robotic Grasp Optimization from Contact Force Analysis.

\end{IEEEbiography}

\begin{IEEEbiography}[{\includegraphics[width=1in,height=1.25in,clip,keepaspectratio]{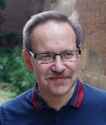}}]{Benoni B. Edin} is a professor of physiology at The Department of Integrative Medical Biology, Ume\r{a} University, Sweden. He earned his MD in 1983 and PhD in 1988. After spending 2 year as a post-doctoral fellow with Prof. James H. Abbs at \emph{The Neurology Department}, University of Wisconsin, Madison, WI, he returned to Sweden. Taking advantage of neurophysiological and behavioral methodologies, including microneurography, he has focussed on issues in the field of somatosensory mechanisms and their role for motor control.

\end{IEEEbiography}

\begin{IEEEbiography}[{\includegraphics[width=1in,height=1.25in,clip,keepaspectratio]{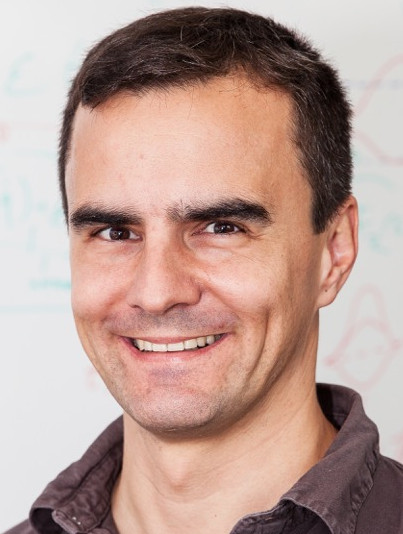}}]{Jan Peters}
is a full professor (W3) for Intelligent Autonomous Systems at the Computer Science Department of the Technische Universitaet Darmstadt and at the same time a senior research
scientist and group leader at the Max-Planck Institute for Intelligent Systems, where he heads the interdepartmental Robot Learning Group. Jan Peters has received the Dick Volz
Best 2007 US PhD Thesis Runner-Up Award, the 2012 Robotics: Science \& Systems - Early Career Spotlight, the 2013 INNS Young Investigator Award, and the IEEE Robotics
\& Automation Society‘s 2013 Early Career Award. In 2015, he was awarded an ERC Starting Grant. Jan Peters has studied Computer Science, Electrical, Mechanical and Control
Engineering at TU Munich and FernUni Hagen in Germany, at the National University of Singapore (NUS) and the University of Southern California (USC), and been an visiting
researcher at the ATR Telecommunications Research Center in Japan. He has received four Master‘s degrees in these disciplines as well as a Computer Science PhD from USC.
\end{IEEEbiography}

\end{document}

%% file: abstract.tex
\begin{abstract}

Grip control during robotic in-hand manipulation is usually modeled as part of a monolithic task, relying on complex controllers specialized for specific situations. Such approaches do not generalize well and are difficult to apply to novel manipulation tasks. Here, we propose a modular object stabilization method based on a proposition that explains how humans achieve grasp stability. In this biomimetic approach, independent tactile grip stabilization controllers ensure that slip does not occur locally at the engaged robot fingers. Such local slip is predicted from the tactile signals of each fingertip sensor i.e., BioTac and BioTac SP by Syntouch. We show that stable grasps emerge without any form of central communication when such independent controllers are engaged in the control of multi-digit robotic hands. These grasps are resistant to external perturbations while being capable of stabilizing a large variety of objects.
\end{abstract}

%% file: Introduction.tex
\section{Introduction}
\label{sec:intro}

Robotic grasping and in-hand manipulation are traditionally viewed as monolithic planning and control problems. As such, control policies determine the approach strategy and finger placement (contact forces and contact locations) for the entire hand, while considering finger trajectories, force and contact profiles throughout the entire manipulation task~\cite{hertkorn2013planning}. This monolithic formalization requires accurate kinematic and dynamic models of the hand and object along with precise sensing of hand and object position as well as interaction forces. In practice, however, control eventually becomes largely data-driven as such models are rarely available and due to the  uncertainty associated with the joint use of all these models~\cite{bohg2014data}.

Data-driven approaches do not come for free. They either require large training data sets~\cite{pinto2016supersizing,yahya2017collective,calandra2017feeling}, restrict the tasks to sufficiently similar scenarios~\cite{bohg2014data,van2015learning}, or rely on low-dimensional representations such as synergies~\cite{prattichizzo2013motion} and motion primitives~\cite{kazemi2013robust}, that encode the considered manipulation task. Consequently, learned polices inherently couple the employed degrees of freedom, resulting in solutions that are task- and platform-specific. Furthermore, incorporating tactile feedback from all fingers into a control policy quickly becomes intractable given the dimensionality of the feedback signals. In short, low-level control policies that both deal with uncertainty (e.g., in contact locations and forces) and generalize well beyond a limited set of cases, need to be both data-driven and modular.

\begin{figure}[t!]
    \centering
    \includegraphics[width=\linewidth]{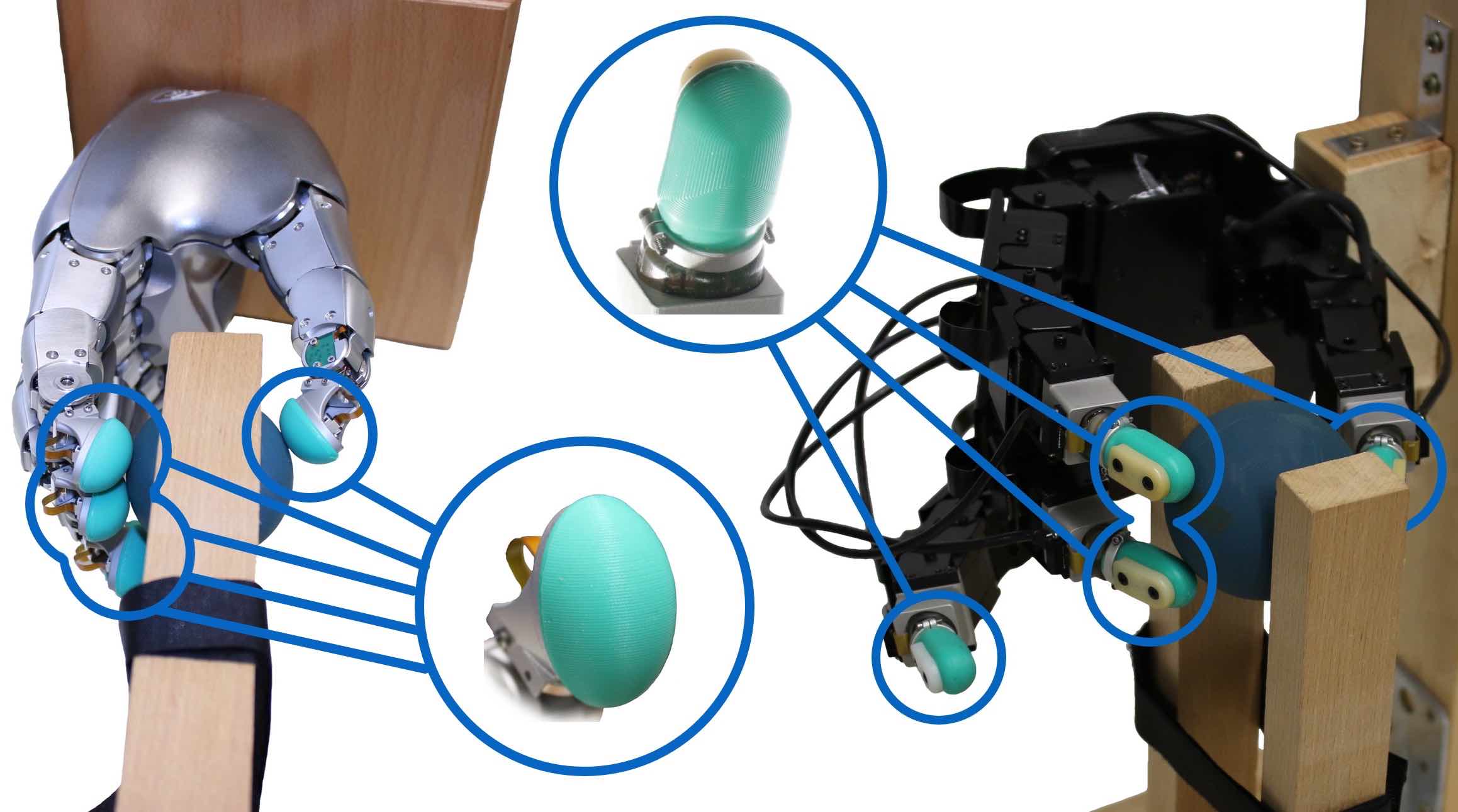}
    \caption{\label{fig:our:hands}The proposed independent finger grip stabilization controller was successfully evaluated on the four finger Allegro Hand (right) and on the five finger Wessling Hand (left). The fingertips of both hands are equipped with Syntouch's BioTac or Biotac SP sensors, respectively.}
\end{figure}

Ensuring grip stability is central to both stabilizing an object in the hand and moving an object between stable grip configurations.
Classical robotics approaches often rely on measures such as form- or force-closure for assessing grip stability -- but with imperfect models and contact/force sensing, using such measures is very challenging.
As a result, many researchers have proposed alternative grasp stability measures~\cite{bekiroglu2011assessing, dang2014stable, madry2014st, li2014learning, romano2011human} and developed accompanying control strategies.

In contrast, human grasping and manipulation appears to be largely data-driven~\cite{Johansson2009} despite relying on feedback signals of huge dimensionality and relatively low precision when control compared to robots. As deduced from several behavioral studies\cite{edin1992independent,burstedt1997coordination,flanagan1999control,manis2015independent}, human grasp control strategies seem to be modular and based on local sensing and actuation, rendering the control of the fingers largely independent from each other, i.e., Independent Finger Control~\cite{edin1992independent}. Specific grasps and force distributions appear to emerge from tactile feedback as the fingers interact with objects. Clearly, such an approach would be desirable for robotic grasping and manipulation.

Inspired by progression from one-finger over two-fingers to the whole hand proposed by~\cite{Klatzky1987object} in the context of tactile object exploration, by early studies of grasp stability using tactile feedback~\cite{bicchi1989augmentation} and by the independent control hypothesis in human grip control by~\cite{edin1992independent,burstedt1997coordination}, we have developed independent control policies based on tactile feedback for each finger that in conjunction generalize from one-finger to five-finger gripping and in-hand manipulation.

To achieve this, we equipped the robotic fingertips of two hands with multimodal fingertip sensors (BioTac and BioTac SP for the four finger Allegro and five finger Wessling Hand, respectively; Figure~\ref{fig:our:hands}), each with a learned predictive model of future slips based on the tactile feedback acquired during finger-object interactions. The local control laws in each finger counteract future slips, ideally preventing them. The resulting control law is capable of stabilizing objects against other objects (such as a table or a wall), jointly stabilizing objects with more robotic fingers (as in in-hand object stabilization or gripping) or against the hand of a human operator (human-robot joint stabilization). It can also be employed for in-hand manipulation by stabilizing an object with several fingers while one or more fingers move the object within the stable grip. The coordination between modular finger controller occurs only indirectly through the tactile signals observed by each finger.

This modular approach enables a higher-level planning system to operate with less object knowledge while requiring simpler models for control than analytical approaches. In contrast to monolithic data-driven approaches, the proposed reactive control framework therefore can be expected to generalize across multiple tasks, a variety of objects and different robotic platforms.
Our approach reproduces findings in human motor control where the absolute amount of force applied by single digits will always settle just above the minimal amount of forces to prevent slip~\cite{Johansson2009,edin1992independent} in static setting.

%% file: PredictingSlip.tex
\section{Modular Tactile Sensing-based \\ In-Hand Object Stabilization}
\label{sec:modular:tactile:sensing:based:in:hand:object:stabilization}
\begin{figure}[b]
	\centering
	\includegraphics[width=\linewidth]{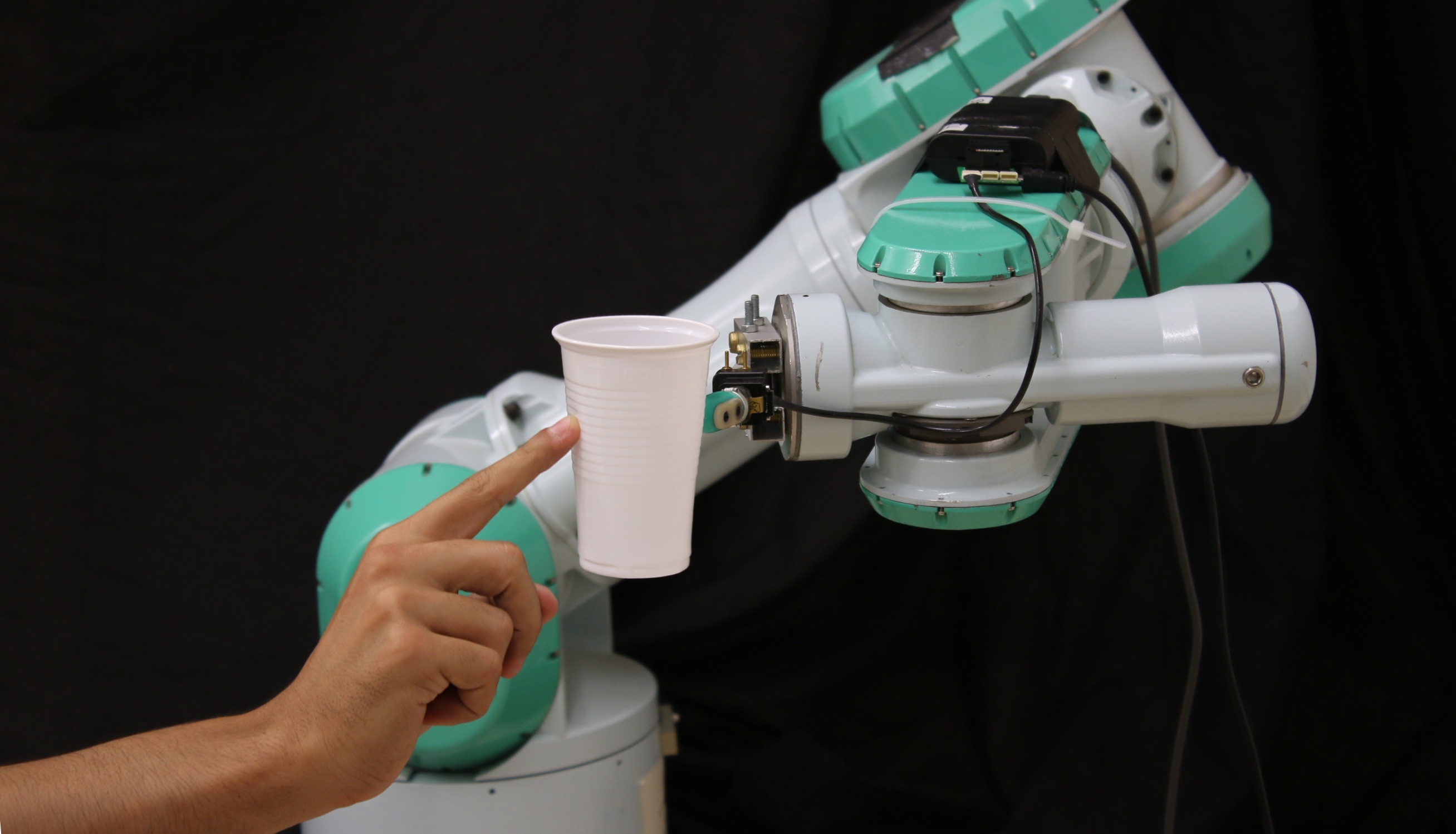}
	\caption{\label{fig:single:finger:stabilization}
		\emph{Single finger stabilization}. As described in Sec.~\ref{sec:single:finger:stabilization}) single finger stabilization was decomposed into a slip prediction component and a control law that attempts to obtain the minimum force where no slip occurs. Slip predictors trained by pinning objects against a fixed plane readily generalized to multi-finger scenarios. Here, we depict a PA10 with a finger equipped with a single BioTac sensor that after training successfully stabilizes an object together with a human finger. The same predictor and controller were used in the multi-finger scenarios through-out this paper.}
\end{figure}

As foundation, this section introduces a single finger tactile control approach which works well for stabilizing objects pinned against other objects. Subsequently, we discuss how this approach can be used in multi-finger settings, i.e., fingers of one robot or those of several agents; see Fig.~\ref{fig:single:finger:stabilization} for an experiment where this method stabilized an object jointly with a human finger.

\subsection{Single-Finger Stabilization based on Slip Prediction \label{sec:single:finger:stabilization}}

Human ability to perceptually discriminate forces applied by their fingertips is limited (Weber fractions typically 5-10\%, \cite{pang1991manual}). Accordingly, tactile information other than those directly related to fingertip force or pressure seem to be employed in human force adjustment strategies during object grasping. As slipping is directly connected to the stability of the interaction with the environment, it is considered crucial for human manipulation~\cite{Johansson2009}. Thus, for single finger tactile stabilization of objects, we recognize two necessary components \cite{edin1992independent}: (1)~a slip predictor and (2)~a force adjustment method based on slip prediction.

\subsubsection{Slip Prediction} \label{sec:slip-prediction}

is formulated as a classification problem ~\cite{veigastabilizing}, relying on a classifier \(f(\cdot)\) that predicts the slip state at time $t+\tau_{f}$ , with $\tau_{f}$ representing the prediction horizon (here $\tau_{f} = 10$, with 10 time steps corresponding to 100 ms). To this end, features $\phi(\cdot)$ of the raw sensor signals $\mathbf{x}_{t}$ (here a vector $[\mathbf{x}_{t}, \Delta x]$ where $\Delta x = \mathbf{x}_{t} - \mathbf{x}_{t-1}$) were extracted for a time window $T = (t-\tau_H):t$, where $\tau_H$ is the tactile history (here $\tau_H=1$). The slip predictor, i.e., $f(\phi(x_t)$, was trained to correctly label the slip state, $c$, at time $t+\tau_f$
\begin{equation}
c_{t+\tau_{f}} = f(\phi(\mathbf{x}_{(t-\tau_H):t}))
\label{eq:prediction model}
\end{equation}
as one of the classes in the set $c_{t+\tau_{f}} \in \{\texttt{slip}, \texttt{contact}, \neg \texttt{contact} \}$. For an in-depth study of how the feature function affects the detection and prediction of slip, the reader is referred to our previous work ~\cite{veigastabilizing}.

\subsubsection{Force adjustment} \label{sec:force-adjustment}

is accomplished through a control law that converts the predicted slip state, $c$, at time $t+\tau_f$ into adjustments in the applied normal force. Most robotic hands are controlled in joint or end-effector positions rather than applied forces. To make the controller applicable across a range of robotic hands, our controller therefore adjusted the desired task space velocities, $\mathbf{\dot{s}}_{t}$, rather than controlling force explicitly. Hence, whenever \texttt{slip} was predicted, we increased the normal force, $F_N$, alternatively slowly decreasing the force while keeping the object stable, in line what has empirically been found during human grasping. This behavior was achieved by using a leaky integrator
\begin{equation}
y_t =  \alpha y_{t-1} + ( 1-\alpha ) L
\label{eq:leaky integrator}
\end{equation}
to control the task space velocity in the contact normal direction, i.e.,
\begin{equation}
\mathbf{s}_t = \mathbf{N}_t y_t.
\label{eq:task-velocity}
\end{equation}
Here, $\alpha$ is the leakage at each time step and $\mathbf{N}_t$ is a unit vector pointing in the contact normal direction. The integrator input signal $L$ changes with the predicted contact state $c_{t+\tau_{f}}$, increasing the accumulated response when \texttt{slip} is predicted and allowing the integrator to leak if \texttt{contact} is predicted, i.e.,
\begin{equation}
	L = \begin{cases}
		1 		& \mbox{if}\; c_{t+\tau_{f}} = \texttt{slip}, \\
		0  					 	&  \textrm{otherwise}
	\end{cases}
\end{equation}
This integrator thus operated as a smoothing filter which was important given the discrete nature of the slip predictor outputs. In multi-fingered scenarios, any oscillations in the controller response would propagate to other fingers engaged in the grasp and cause instability. While still allowing the fingers to react to all oscillations, the integrators manage the intensity of the response, slightly changing the applied force for instantaneous perturbations or greatly increasing the applied force for more persistent perturbations.

Finally, a minimum integrator response $y_{min}$ is required to avoid oscillations around low integrator responses values where slip is imminent. However, instead of specifying $y_{min}$, each finger estimates its minimum response by observing the first slip transient following a first stable period. The minimum response is then defined as the the response $y_{t}$ where the first transition from \texttt{contact} to \texttt{slip} occurs
\begin{equation}
y_{min} = y_{t}\; \mbox{if} \; \Delta c =  \texttt{contact} \rightarrow \texttt{slip}.
\end{equation}
This minimum response implicitly defines the minimum fingertip normal force necessary to prevent slips and makes the controller responsive to the prevailing friction at its digit-object interface.

\subsection{Multi-Finger Gripping by Single-Finger Slip Control}
\label{sec:multi:finger:scenario}
When progressing towards in-hand stabilization and in-hand manipulation, more fingers are required and the complexity of the tasks quickly scales accordingly with hand dexterity. Generally, a higher dimensionality can be coped with either by identifying a lower-dimensional manifold for the problem or by decomposing the problem. Following the core insight in~\cite{edin1992independent} that human multi-finger grip stabilization appears to be accomplished by separate neural circuits that interact through the object instead of via the central nervous system, we hypothesize that \textsl{multi-finger robot gripping can be accomplished using the same single-finger stabilization controller on each finger independently}. As a first scenario, we reproduce the scenario from~\cite{edin1992independent}, where two humans jointly hold an object using one finger each with the same apparent ease as if a single person use the index finger and thumb of one hand or one finger from both hands. The underlying neural control appeared to be unaffected by the precise task condition. We reproduce this experiment in a human-robot joint stabilization task as shown in Fig.~\ref{fig:single:finger:stabilization}: the single finger controllers of both the human and the robot worked well together without any precautions.

To fully utilize the dexterous capabilities of the hand, we propose that each hand should be considered a set of independently controlled fingers pertaining specifically to stabilization. This independent control approach obviously still requires a higher level planning approach for fingertip placement, as well as making and breaking of contact with the object.


A set of independent fingers -- in contrast to a fully connected manipulator -- allows decomposing the object stabilization control problem such that each finger separately predicts future slip based on tactile sensing, counteracting it by independently adjusting the applied forces. While synchronization only through the tactile feedback may appear counter-intuitive, it actually greatly reduces the dimensionality of the control problem while ensuring that the fingers affect each other only when necessary for object stabilization. As a result, it not only becomes more straightforward to design stabilizing control laws but the synchronization becomes more robust.

%% file: ExperimentalEvaluation.tex
\section{Experimental Evaluation}
\label{sec:exp-eval}

The proposed independent finger control law (from Sec.~\ref{sec:modular:tactile:sensing:based:in:hand:object:stabilization}) is evaluated both to constructively verify the independent finger control hypothesis as well as to show that the proposed approach works sufficiently well in practice. We begin by stabilizing several objects with a varying number of fingers, using the Allegro hand, without any external perturbations (Sec.~\ref{sec:multi-finger-grip-stabilization}), and demonstrate that a control strategy working under the proposed hypothesis is able to re-stabilize objects in-hand throughout sequences of externally applied perturbations (Sec.~\ref{sec:grip-stabilization-under-external-perturbations}). The presentation of the results is preceded by a detailed description of the experimental setup, i.e., robotic platform and an account of the tactile sensors mounted on the platform as well as the sensors used to measure the external perturbations (Sec.~\ref{sec:testing-platforms}), and a detailed outline of the procedure used to acquire the ground truth data for the slip classifiers (Sec.~\ref{sec:tactile-training-data}).

\subsection{Experimental Setup: Testing Platform \& Tactile Sensors}
\label{sec:testing-platforms}
To demonstrate the independent finger control, the control scheme was implemented on two robotic hands: The four finger Allegro Hand and the five finger Wessling Robotic Hand.

The Allegro Hand (Wonik Robotics GmbH, \url{www.simlab.co.kr}; Fig.~\ref{fig:our:hands}), is a lightweight four fingered hand with four joints per finger, for a total of 16 actuated degrees of freedom. The thumb has an abduction joint, two metacarpal joints (rotation and flexing) and a proximal joint. The remaining fingers do not have abduction joints and instead have a distal joint. A PD controller was used to control the robot joint positions. One end-effector was defined for each fingertip and their positions were controlled by estimating the desired joint velocities, by means of the Jacobian Pseudo-Inverse, and integrating the estimations to acquire the desired joint positions.

The Wessling Robotic Hand has five modular fingers, each with four joints where two of these four joints are coupled and cannot be moved independently (Wessling Robotics, \url{www.wessling-robotics.de}; Fig.~\ref{fig:our:hands}). A PD controller is used for joint position control and a Pseudo-Inverse Jacobian controller is used for controlling the end-effector position of each finger. The control signals are sent to a real-time machine where the conversion to torque is performed by an joint impedance controller from Wessling Robotics~\cite{chen2011experimental}.

While the Allegro Hand has one finger fewer than the Wessling Robotic Hand, it is more compliant and its workspace is larger than that of the Wessling Hand. The base control loops of each hand operate at different frequencies, i.e., 300~Hz and 1~kHz for the Allegro and Wessling Hand, respectively. However, despite these differences, the slip prediction based controllers were the same, each controller trained on data from the respective fingertip sensors.

BioTac and the BioTac SP tactile sensors (SynTouch Inc., \url{www.syntouchinc.com}; Fig.~\ref{fig:our:hands}) were mounted on the Allegro and Wessling Hand, respectively, and served as fingertips. Both provide multi-modal responses composed of low and high frequency pressure ($P_{\textrm{dc}}$ and $P_{\textrm{ac}}$), local skin deformations ($E$), temperature and thermal flow ($T_{\textrm{dc}}$ and $T_{\textrm{ac}}$). The sensor consists of a conductive fluid captured between a pliable skin and a rigid core. The core surface is covered with impedance sensing electrodes (19 for BioTac; 24 for BioTac SP). The pressure signals are acquired by a pressure transducer, skin deformation is measured through local impedance changes measured by the electrodes and temperature is regulated by a thermistor. All data channels of the sensor are sampled at a rate of 100~Hz. The high frequency pressure signal is acquired internally by the sensor at a rate of 2.2~kHz, but is available for readout at 100~Hz, producing batches of $22$ values every 10~ms. Considering all channels and the Pac batch data, the sensors output a total of 44 or 49 values every 10~ms.

Finally, the Optoforce OMD-D20 3D (Optoforce Ltd., \url{www.optoforce.com}) is an optical force sensor (insets of Fig.~\ref{fig:optoforce:vs:leak}) that was used to measure the magnitude of external perturbations applied on the objects during in-hand re-stabilization experiments. The Optoforce reconstructs the magnitude and direction of the applied force from the values of four light sensitive photo-diodes that detect the amount of reflected light by interior surface diodes. The sensor has a nominal sample rate of 100~Hz.

\begin{figure}[t]
	\centering
	\includegraphics[width=\linewidth]{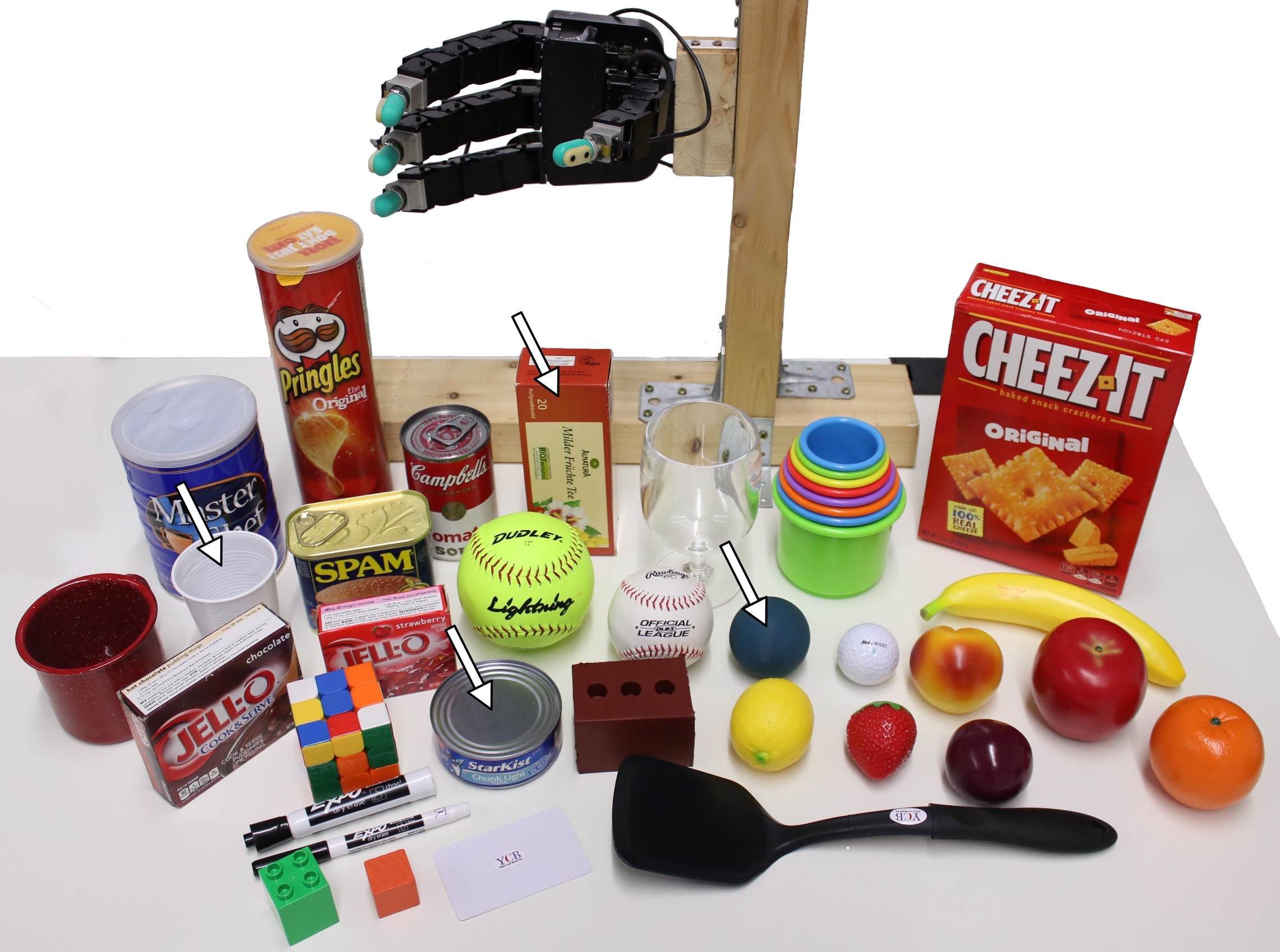}
	\caption{\label{fig:test:object:set}\emph{Test objects}. The majority of the objects were from the YCB object set \cite{calli2015ycb} where only the tea box and the white plastic cup are not in the original set. The \emph{training set} (indicated by the white arrows) included 4 objects only: a tuna can, a plastic cup, a ball, and a tea box.}
\end{figure}

\subsection{Test and Training Objects}
Our set of 38 test objects belonged with two exceptions (a tea box and a plastic cup) to the YCB object set  \cite{calli2015ycb}, shown in Fig.~\ref{fig:test:object:set}. Among the test objects, the weight varied from 10g to more than 400g and grasp width from less than 10 mm to more than 100 mm. Specifically, the plastic cup (cf., Fig.~\ref{fig:single:finger:stabilization}) was included to assess the performance of the control system when faced with highly deformable objects.

Only 4 objects were used during training: a tuna can, a plastic cup, a ball, and a tea box (arrows in Fig.~\ref{fig:test:object:set}). Successful manipulation of \emph{all} test objects thus implied that the method generalized across grasps and object properties.

\subsection{Tactile Training}
\label{sec:tactile-training-data}

As our independent finger stabilizers reacted to slip-based feedback, it was necessary to train the classifiers responsible for slip prediction. This training required data collected on the real system and ground truth labels for the slip events.

\begin{figure}[b!]
	\centering
  \includegraphics[width=0.9\linewidth]{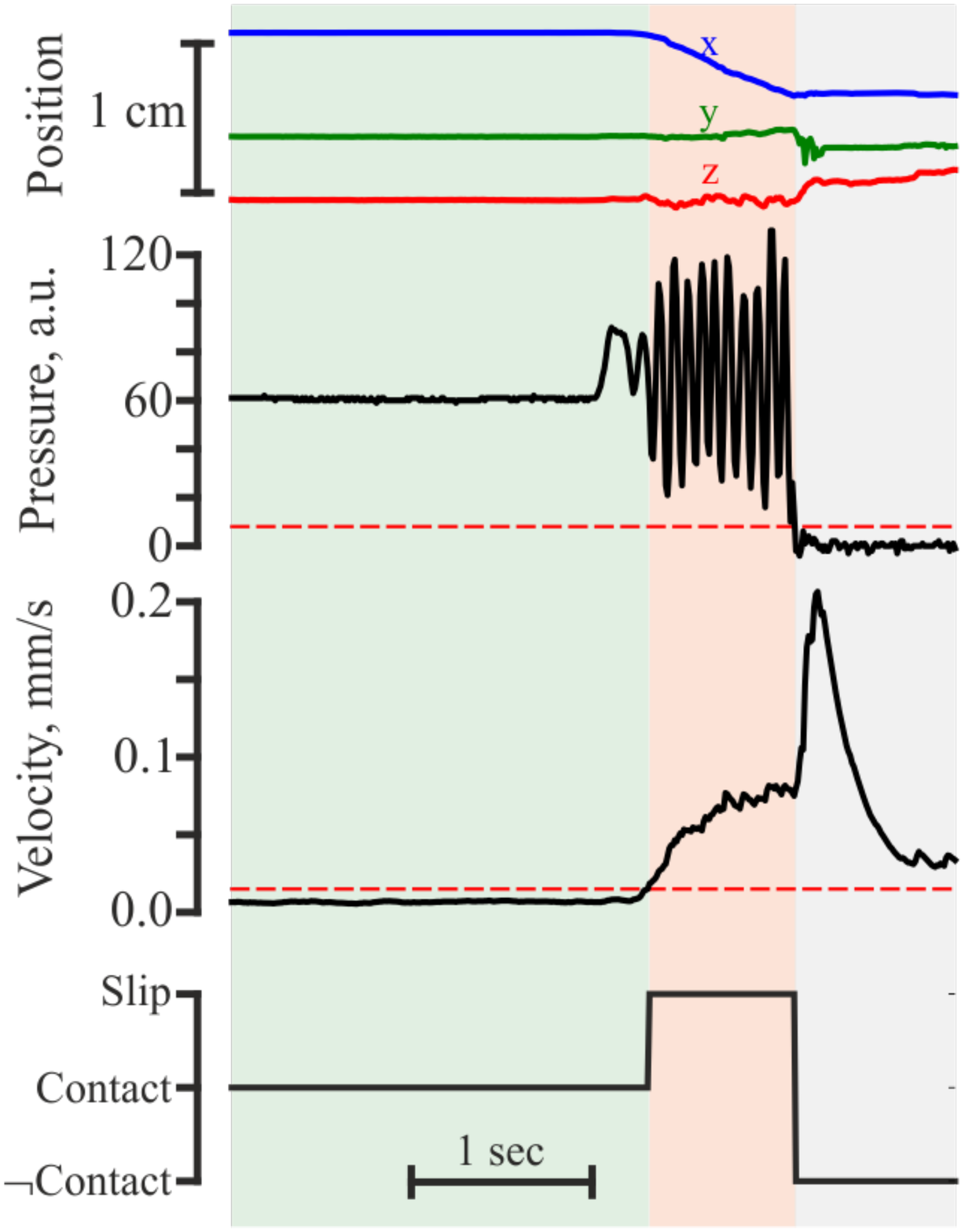}
	\caption{\label{fig:data:slip:labeling} Data from the index finger during a single, representative training trial. The cartesian instantaneous velocity was calculated from differences in finger end-effector position between two consecutive time steps. A pressure threshold, $T_\textrm{Contact}$, and a movement threshold, $T_\textrm{Movement}$, both indicated with red dashed lines, were used to generate the slip ground truth labels shown in the bottom panel.}
\end{figure}

\begin{figure*}[t]
\centering
\includegraphics[width=0.9\textwidth]{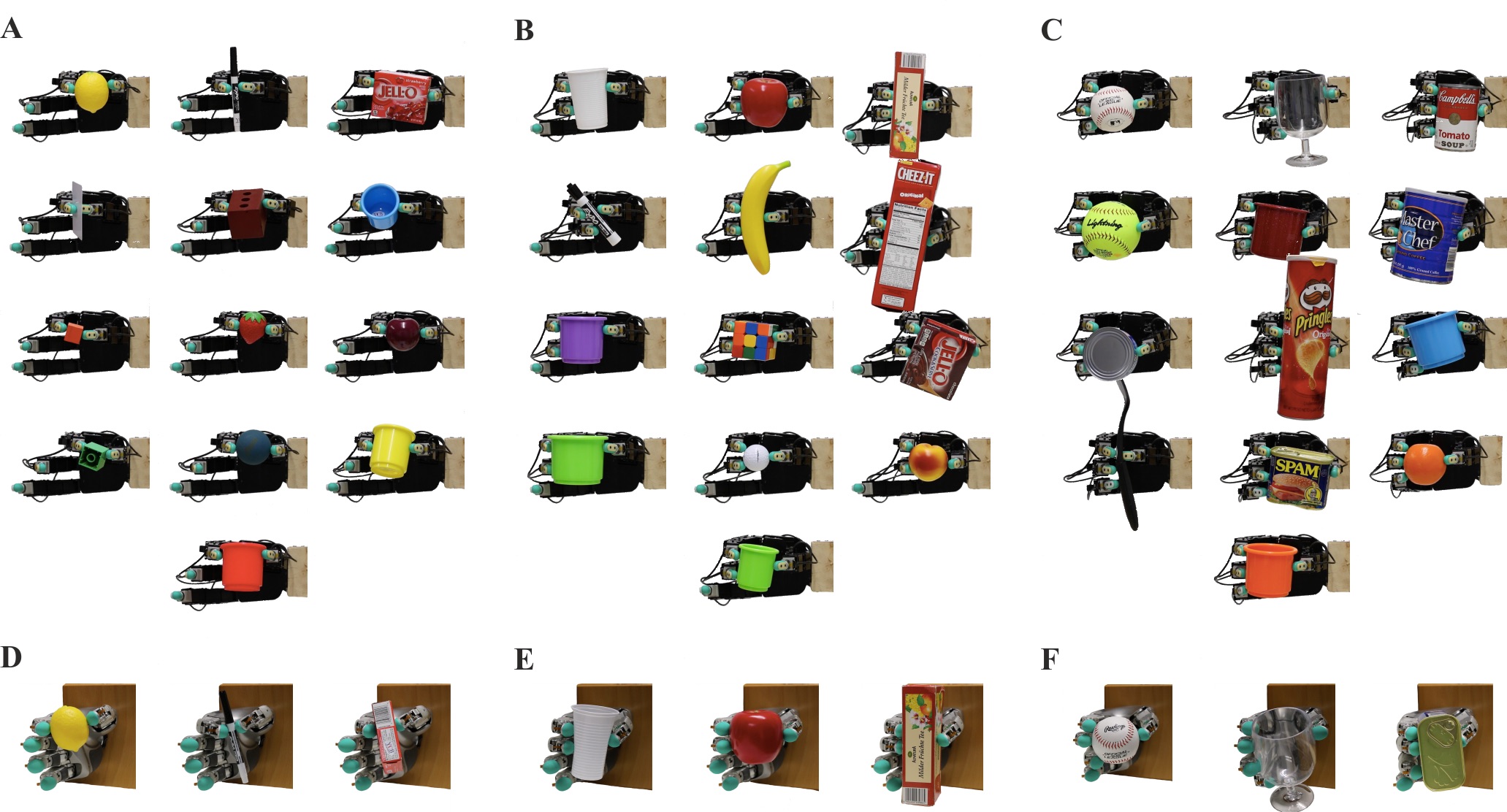}
\caption{\label{fig:grasp:object:set}Stable grasps of a variety of objects. The specific grasp configurations varied from trial-to-trial but always resulted in stable grasps. The panels show (A) two-finger, (B) three-finger and (C) four-finger grasps with the Allegro Hand and (D) two-finger, (E) three-finger and (F) four-finger as well as five-finger grasps with the Wessling Robotic Hand.}
\end{figure*}

To start data collection, one of the training objects was fixated by a support in the hand’s work space (Fig.~\ref{fig:our:hands}). All fingers were positioned in an initial configuration and subsequently flexed until they made contact with the object. Then the pressure applied by each finger was adjusted by a PID controller until a target pressure was reached on each finger. Finally, the fingers moved along the tangential contact plane, surveying the object surface. Acquiring data from three sensors simultaneously reduced the necessary number of training trials. All data from each of the fingers was concatenated into a single data set that was used to train each of the individual slip predictors. The data collection setup is exemplified in Fig.~\ref{fig:our:hands}.

Fig.~\ref{fig:data:slip:labeling} shows a representative, single training trial with data from the index finger. Slip labels were generated automatically from the finger’s end-effector location and the recorded pressure values. The total shift in Cartesian position was calculated from the end-effector position. Since the object was fixated during training, we defined slipping as the state when the finger was in contact (i.e., the recorded pressure was above a certain threshold $T_\textrm{Contact}$) and the finger was moving (i.e., the finger velocity exceeded the movement threshold $T_\textrm{Movement}$; both thresholds are indicated with dashed lines in Fig.~\ref{fig:data:slip:labeling}).

This procedure relied on randomly selected velocities in task space for the object surface surveying. Target pressures were selected from 9 possible values in sensor grounded pressure units: $P~=~[20, 40, 60, 80, 100, 150, 200, 250, 300]$. Spanning the data across multiple pressures in conjunction with randomly selecting the velocity and having distinct contact locations across the three fingers, allowed for training slip classifiers that were not specifically correlated with certain pressures, contact locations or fingertip velocities. In addition, all sensor values concerning pressure or finger deformation were grounded before training, preventing parametric differences in the sensors (for example nominal fluid pressure) from correlating to slip.
Three trials were executed for each value of $P$ on four training objects (Fig.~\ref{fig:test:object:set}) for a total of 108 trials. The resulting data set thus comprised 324 single finger trials across the three engaged fingers and was acquired in less than 15 minutes.

\begin{figure*}[t]
\centering
\includegraphics[width=0.7\textwidth]{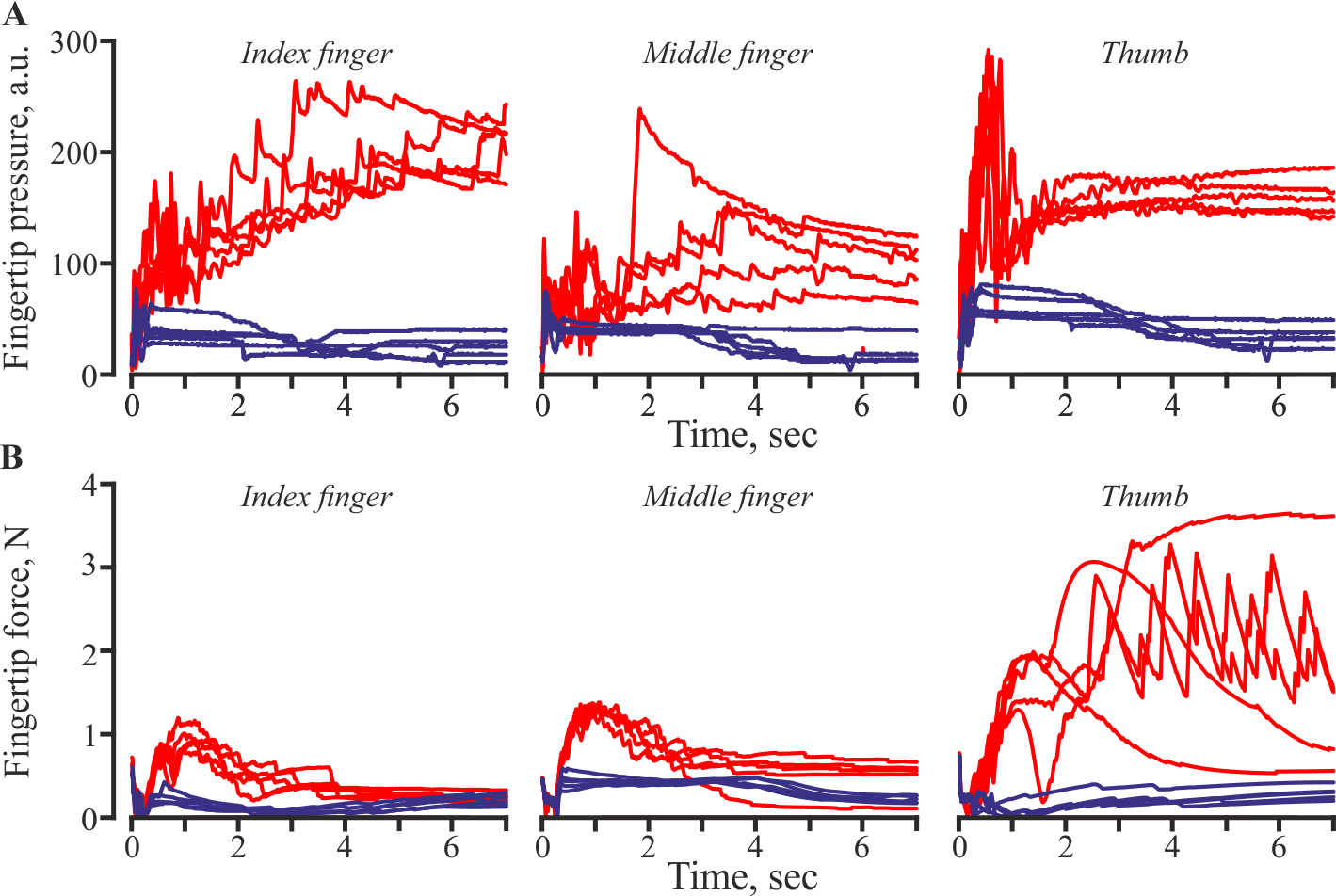}
\caption{\label{fig:pressure:comparison} Pressure and force profiles. A comparatively light object (plastic mug; blue lines) or a heavy object (cracker box; red lines) was grasped five times with the Allegro Hand (A) and the Wessling Robotic Hand (B). While all attempts resulted in stable grasps, the exact configuration varied with the fingertip pressures and forces changing accordingly.}
\end{figure*}

\subsection{Grip Stabilization Evaluation}
\label{sec:grip-stabilization-evaluation}

For the multi-finger grip stabilization scenarios, finger pressure was analyzed and used to make behavioural comparisons across objects (reported in Sec.~\ref{sec:multi-finger-grip-stabilization}). In addition, we assessed the in-hand re-stabilization capability of our approach as the grip was perturbed by an external agent (Sec.~\ref{sec:grip-stabilization-under-external-perturbations}).

Since each finger was controlled independently, the approach was scalable with respect to the number of fingers. In this study, however, we considered grip configurations involving two and three fingers across all test objects (Fig.~\ref{fig:test:object:set}) including the four objects used in the slip predictor training data collection experiments.

\subsubsection{Multi-Finger Grip Stabilization with Independent Finger Control}
\label{sec:multi-finger-grip-stabilization}

To test the validity or our independent finger control hypotheses for grip stabilization, we attempted to stabilize multiple objects with varying number of finger.

We place the robotic hand in an open-hand configuration with an object positioned such that it could be held in an opposition grasp, and then closed two or more fingers (up to four with the Allegro Hand and up to five with the Wessling Robotic Hand). Immediately after all fingers have made contact with the initially supported object, the grip stabilizers were activated and the independent finger stabilization process began, while the object support was removed. To ensure that the object would not be dropped during the activation transient of the grip stabilizers, each controller was initialized to generate a predefined fraction of the maximum output. For deformable objects such as the white plastic cup, this activation resulted in an initial surface deformation that was subsequently automatically reduced.

The control based on independent finger control was able to reliably and consistently stabilize all 39 test objects (Fig.~\ref{fig:grasp:object:set}). For each object and grasp configuration (two-, three- and four-finger grasps with the Allegro Hand and up to five-finger grasps with the Wessling Hand), we recorded five trials each lasting 10 seconds with every object. A grasp was considered stable if the object was not dropped.

Since no desired hand configuration was enforced, the hand adopted slightly different configurations for each object and across repetitions. To study this variability in more detail, we analyzed the grip forces applied by the fingers to different objects. Figure~\ref{fig:pressure:comparison} shows the pressure profiles and estimated forces for the Allegro and Wessling Hand, respectively, for trials with the lightest and one of the heaviest objects, i.e., the white plastic cup and the cracker box. The pressure profiles applied in the Allegro experiments were recorded directly from the BioTac sensors while the estimated forces applied in the Wessling experiments were calculated from joint torques and angles. The data illustrates two important emergent properties of the grasp control. First, finger pressures and forces converged to lower values when gripping the lighter plastic cup than when gripping the cracker box. Second, there was a substantial variability in force sharing between the digits across trials, particularly obvious in the profiles recorded during trials with the cracker box.
Both of these observations can be explained straightforwardly through the design of the controller. Notably, an uncountable number of grip force distributions could result in stable grasps but the control system did not explicitly enforce a specific distribution. Instead, pressure applied by each finger propagated through the object to the other fingers, dynamically impacting the grip force distribution while each controller minimized the risk for local slips keeping the fingertip forces low. The ability to adapt the overall grip force by reactively changing the force applied by each finger contributed to the high generalization capability of our approach, even though no specific object orientation, weight or weight distribution was expected by the stabilizers.

\subsubsection{Grip Stabilization under External Perturbations}
\label{sec:grip-stabilization-under-external-perturbations}

\begin{figure*}[t]
\centering
\includegraphics[width=0.9\textwidth]{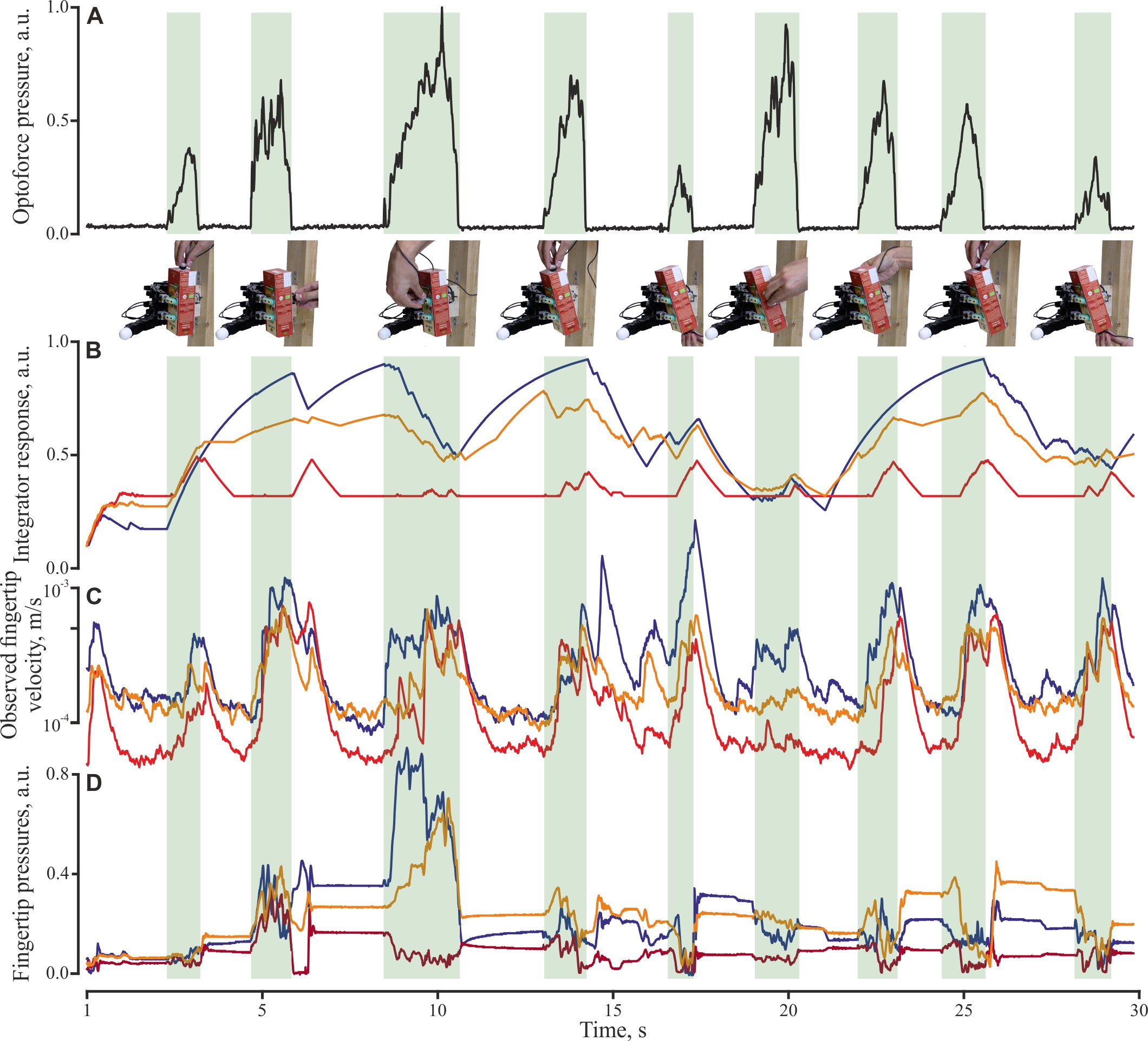}
\caption{\label{fig:optoforce:vs:leak}Responses to external perturbations. The panels show (A) the pressure applied by the experimenter on the surface in the manner shown in the insets, (B) the integrator response of the controller that drives the fingertip velocities (C) the observed fingertip velocities and (D) the applied fingertip pressures by the thumb, index and middle fingers (yellow, blue and red lines). As each controller continuously predicted the contact state 100 ms in the future, the output of their leaking integrators increased whenever a slip was predicted, otherwise allowing the integrator output to decrease slowly to a minimum value. The integrator response determined the necessary fingertip velocity, thereby implicitly managing the applied pressure against the object surface.}
\end{figure*}

To further test the validity of our underlying control hypothesis, we investigated responses to externally applied perturbations (Fig.~\ref{fig:optoforce:vs:leak}). Once the object was stabilized in the robotic hand, the experimenter held an Optoforce sensor and used it to disrupt the object state by applying sequences of irregular disturbances, either to the different surfaces of the objects or to the fingertips, during 30 second recording periods (insets in Fig.~\ref{fig:optoforce:vs:leak}).

For the entire duration of these experiments, the stabilizers invariably counteracted the perturbations successfully by adapting the finger pressures. With every perturbation, we observed a change in the fingertip forces and an increase in the accumulated value of the integrator that regulated the applied velocity. As a result, the individual fingers applied slightly different steady-state forces after each perturbation. For instance, the 1st, 4th and 8th perturbation in Figure~\ref{fig:optoforce:vs:leak} were applied in a similar fashion (i.e., from top) but in response, the independent finger controllers generated different stable grip force distributions. Indeed, while the object was held in a similar position throughout this trial, the pressure distributions across the fingers differed following each perturbation. Changes in fingertip forces due to slip prediction noise or re-stabilization were also frequently observed (e.g., around 16 and 21 second mark).

\subsubsection{Master-Slave Operation}
\label{sec:master-slave-operation}

\begin{figure*}[t]
  \centering
  \subfloat{\includegraphics[height=0.1\textwidth]{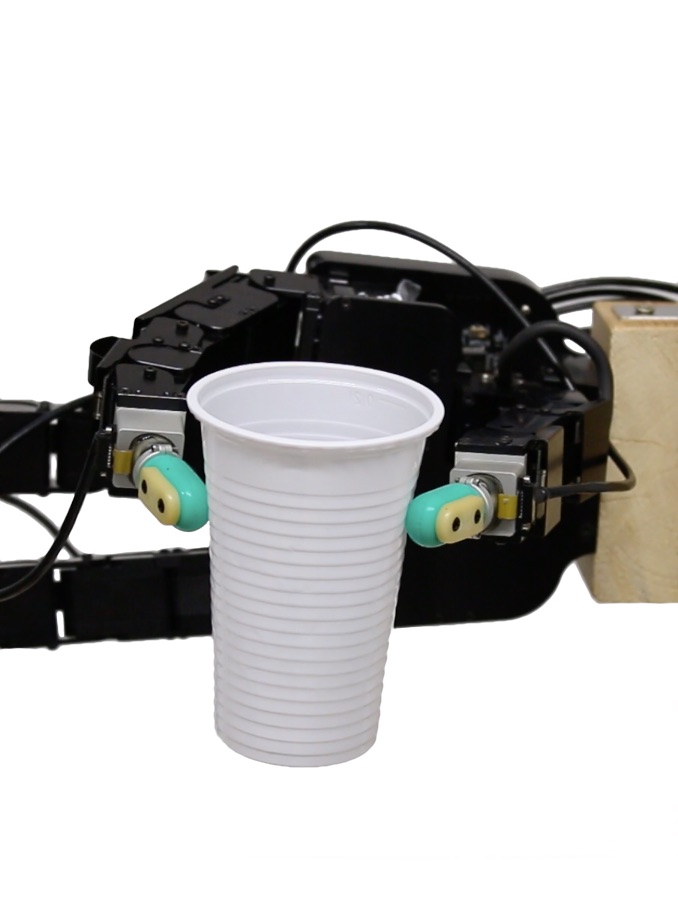}}\hspace{0.001\linewidth}
  \subfloat{\includegraphics[height=0.1\textwidth]{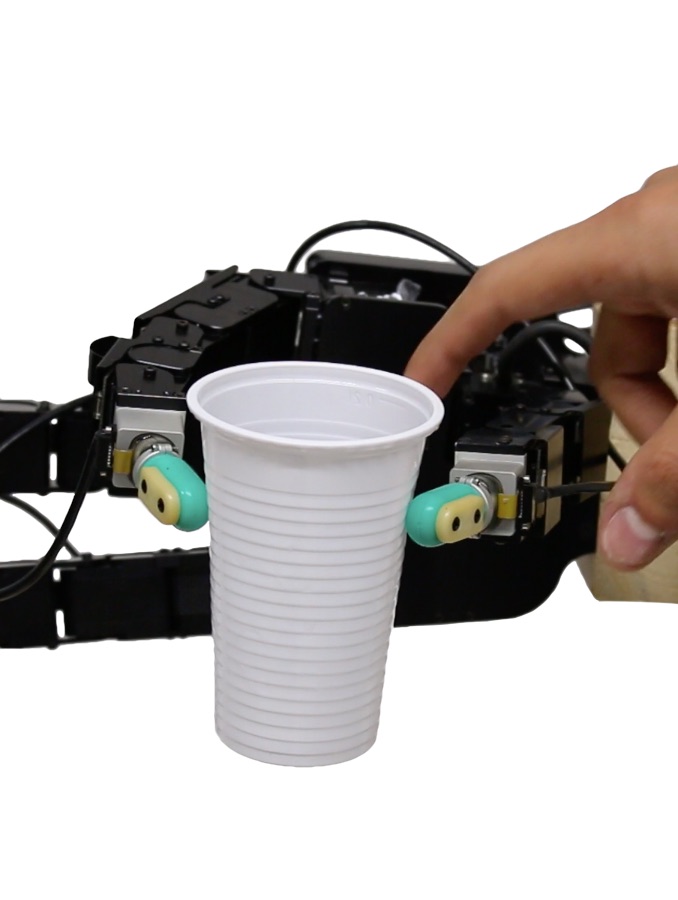}}\hspace{0.001\linewidth}
  \subfloat{\includegraphics[height=0.1\textwidth]{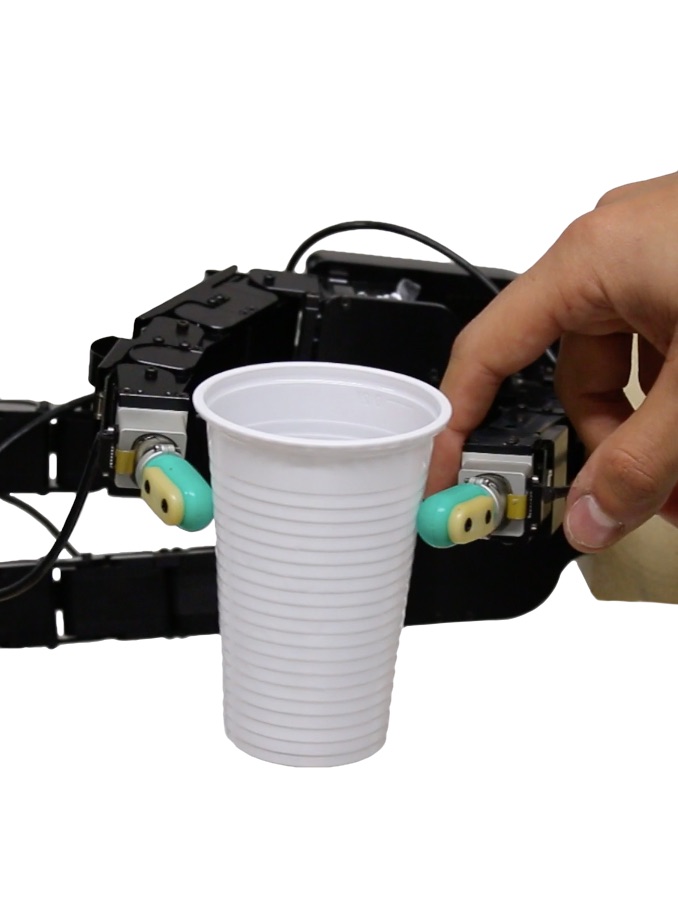}}\hspace{0.001\linewidth}
  \subfloat{\includegraphics[height=0.1\textwidth]{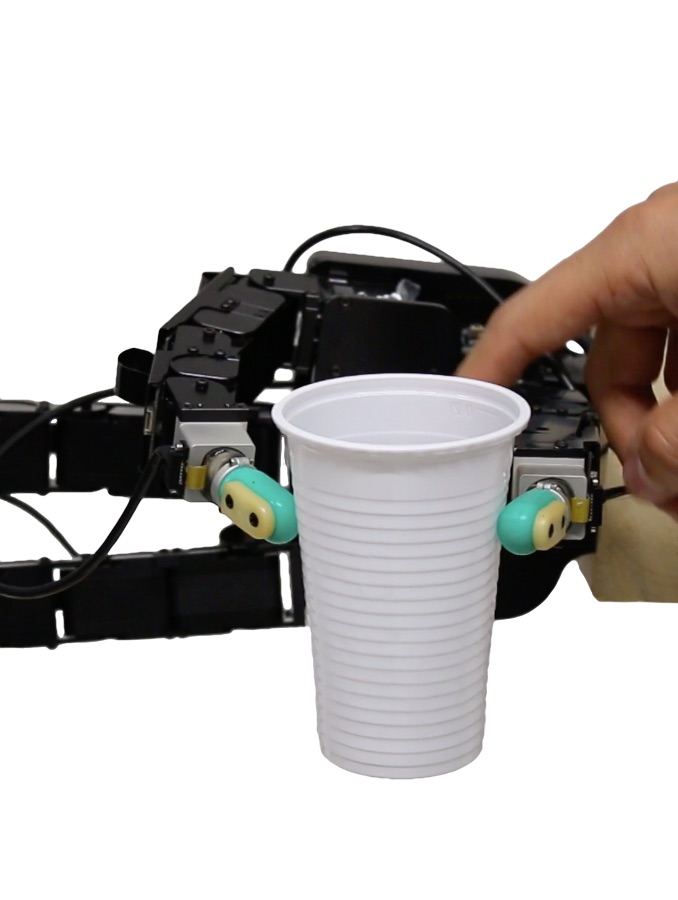}}\hspace{0.001\linewidth}
  \subfloat{\includegraphics[height=0.1\textwidth]{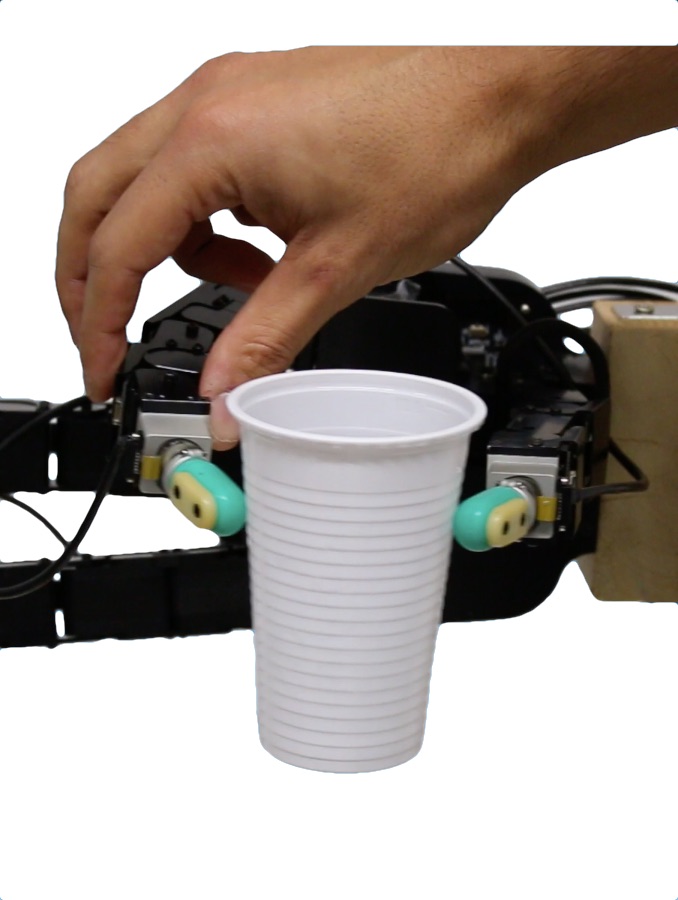}}\hspace{0.001\linewidth}
  \subfloat{\includegraphics[height=0.1\textwidth]{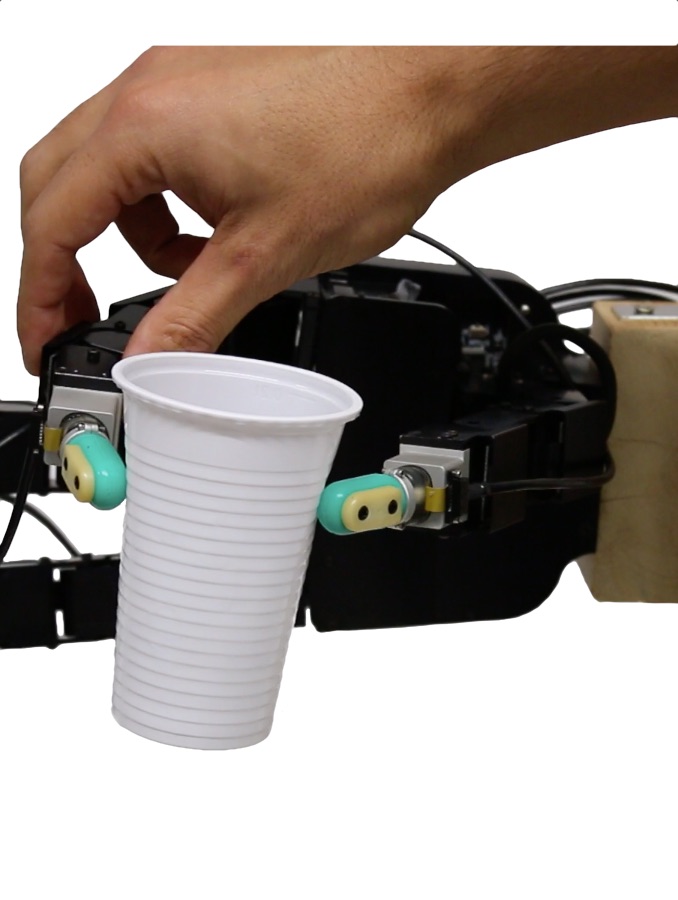}}\hspace{0.001\linewidth}
  \subfloat{\includegraphics[height=0.1\textwidth]{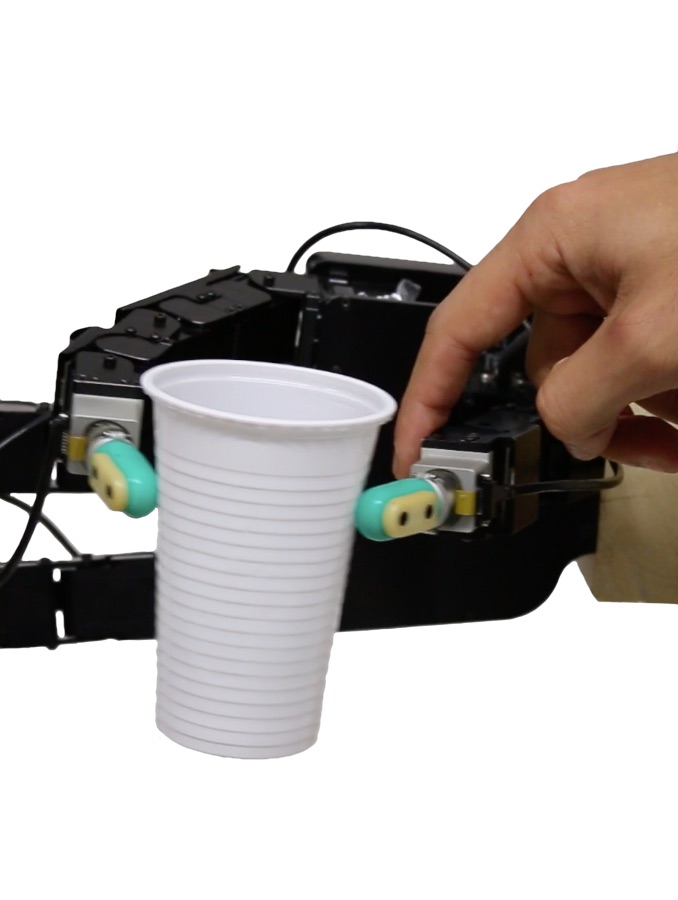}}\hspace{0.001\linewidth}
  \subfloat{\includegraphics[height=0.1\textwidth]{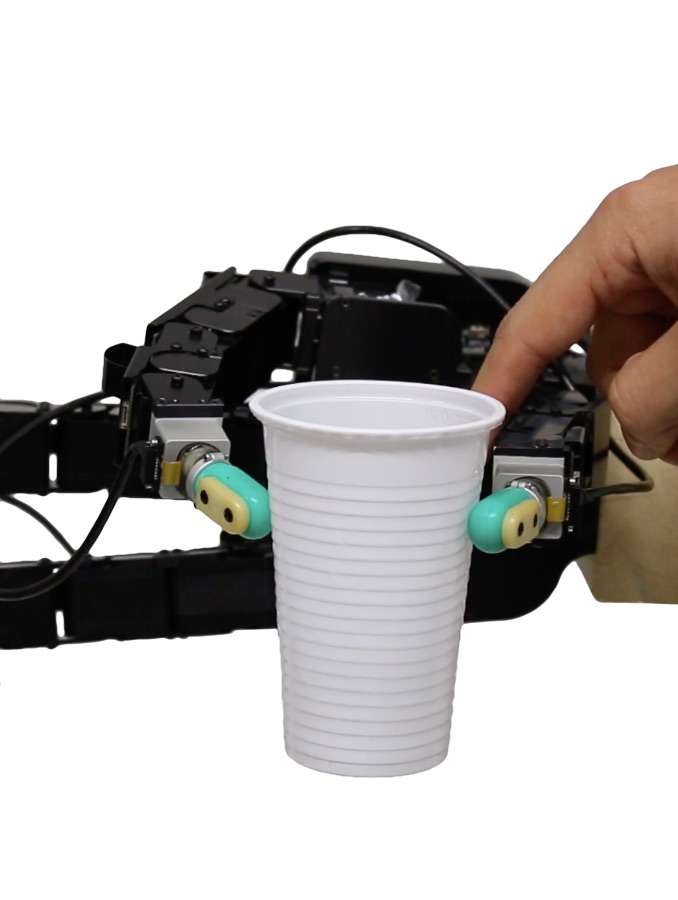}}\hspace{0.001\linewidth}
  \subfloat{\includegraphics[height=0.1\textwidth]{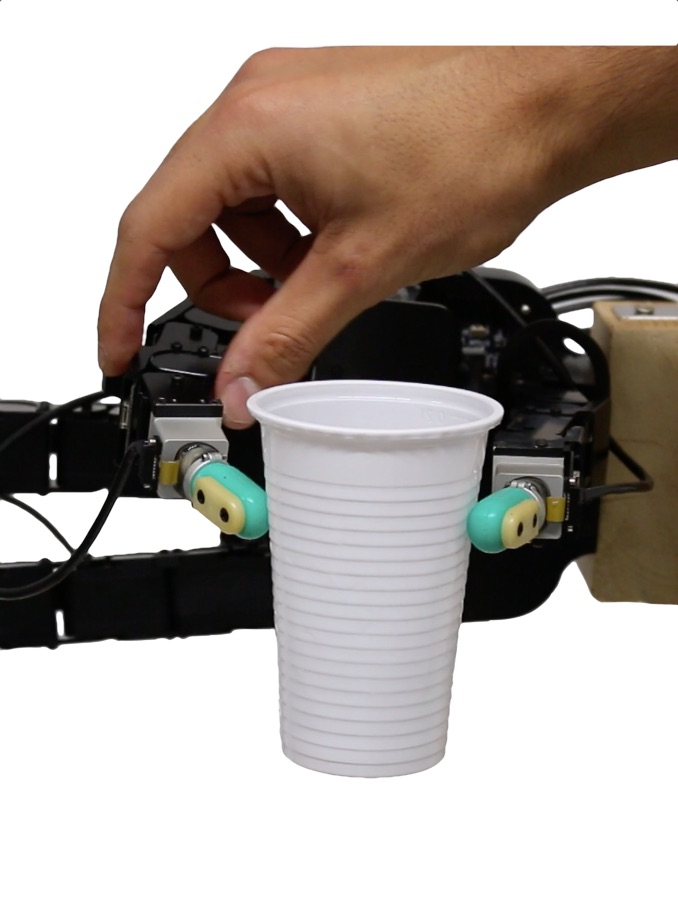}}\hspace{0.001\linewidth}
  \subfloat{\includegraphics[height=0.1\textwidth]{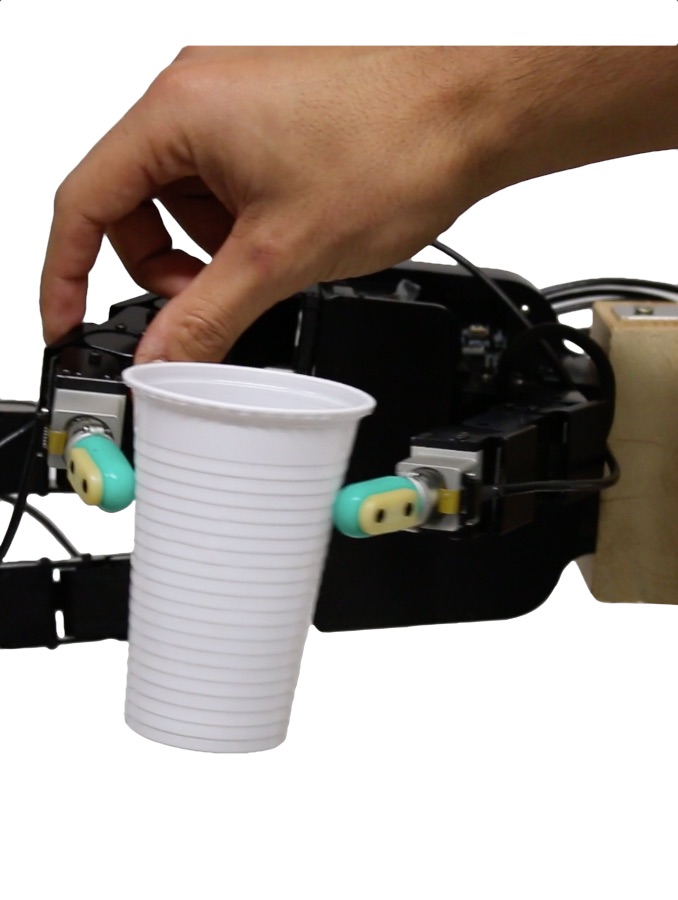}}\hspace{0.001\linewidth}
  \subfloat{\includegraphics[height=0.1\textwidth]{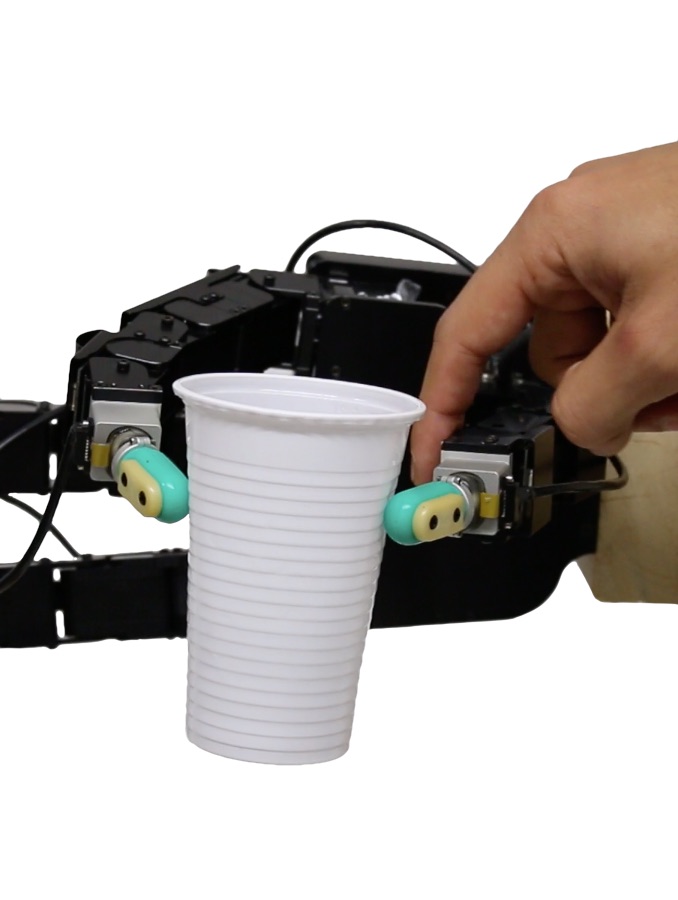}}\hspace{0.001\linewidth}
  \subfloat{\includegraphics[height=0.1\textwidth]{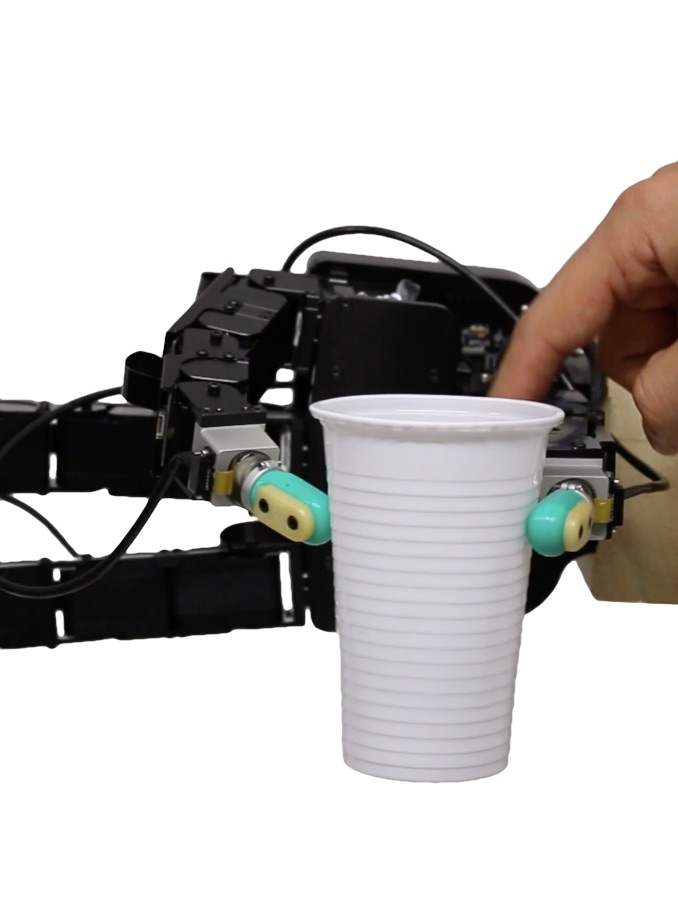}}\\
  \subfloat{\includegraphics[height=0.1\textwidth]{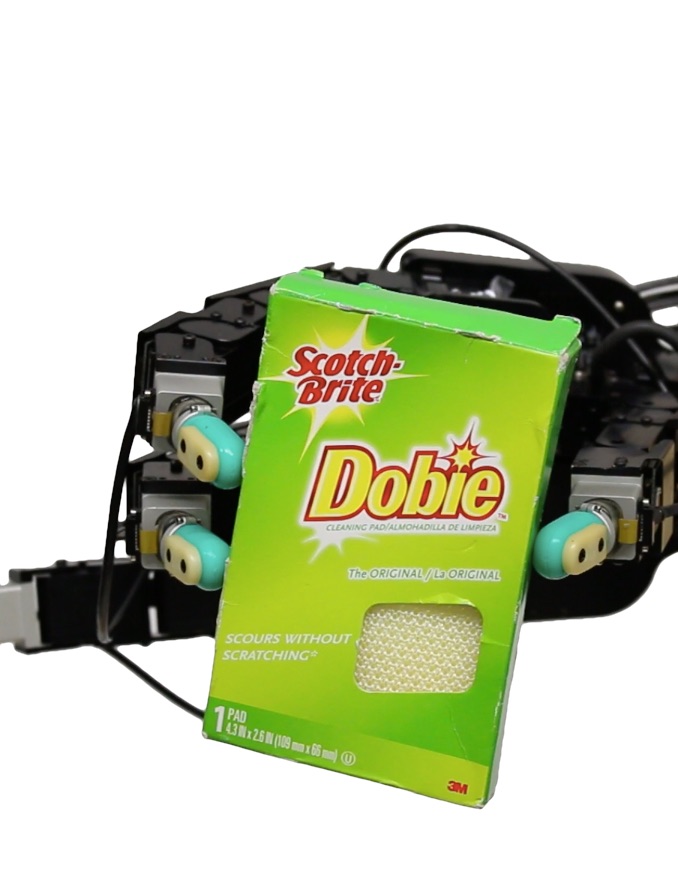}}\hspace{0.001\linewidth}
  \subfloat{\includegraphics[height=0.1\textwidth]{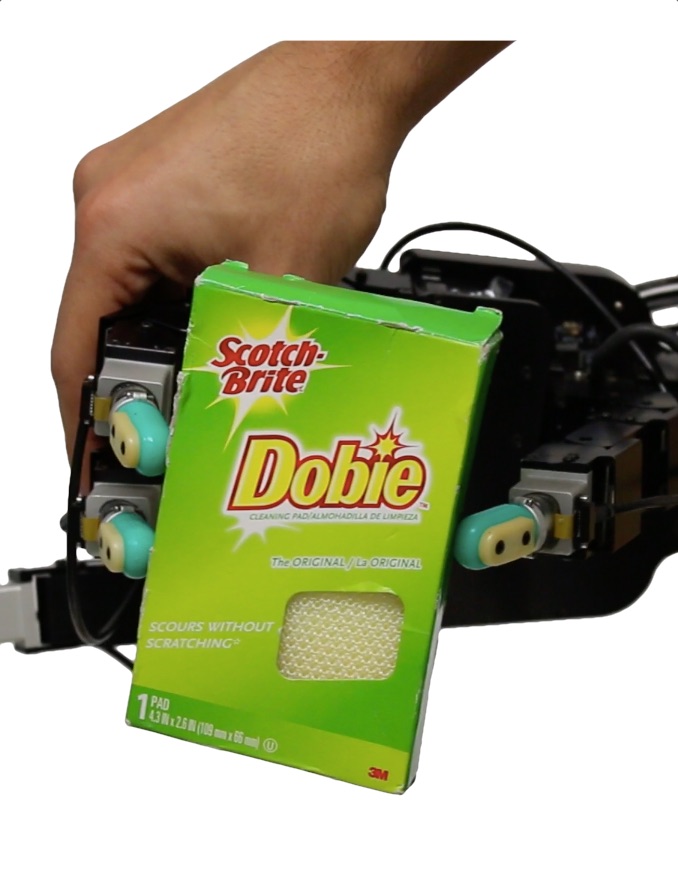}}\hspace{0.001\linewidth}
  \subfloat{\includegraphics[height=0.1\textwidth]{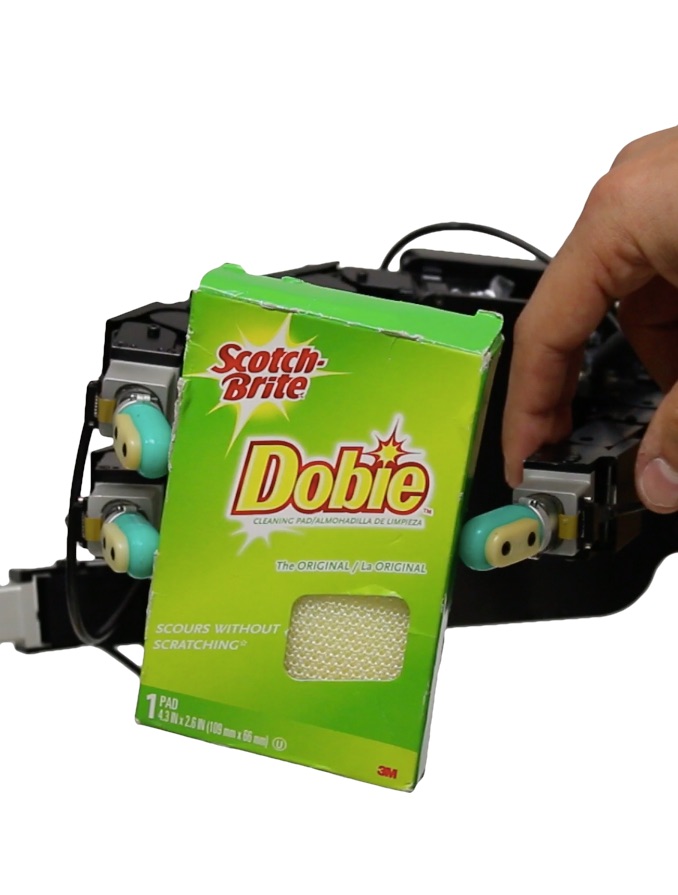}}\hspace{0.001\linewidth}
  \subfloat{\includegraphics[height=0.1\textwidth]{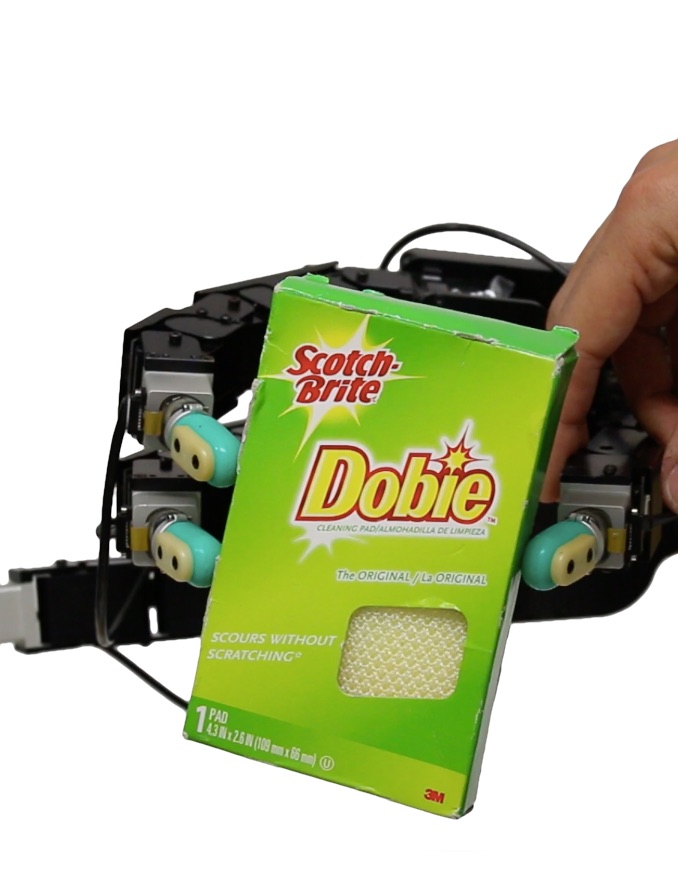}}\hspace{0.001\linewidth}
  \subfloat{\includegraphics[height=0.1\textwidth]{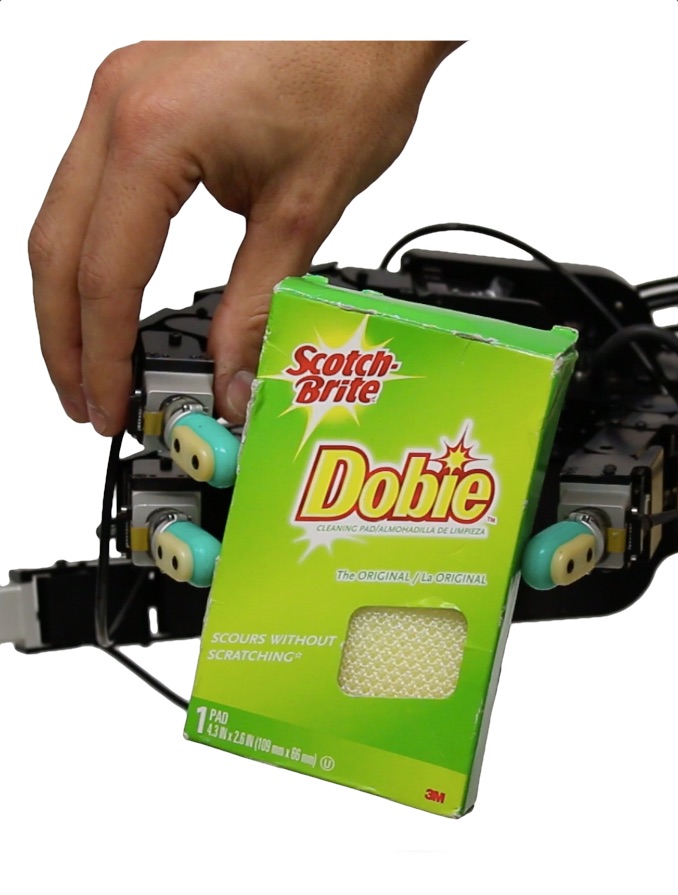}}\hspace{0.001\linewidth}
  \subfloat{\includegraphics[height=0.1\textwidth]{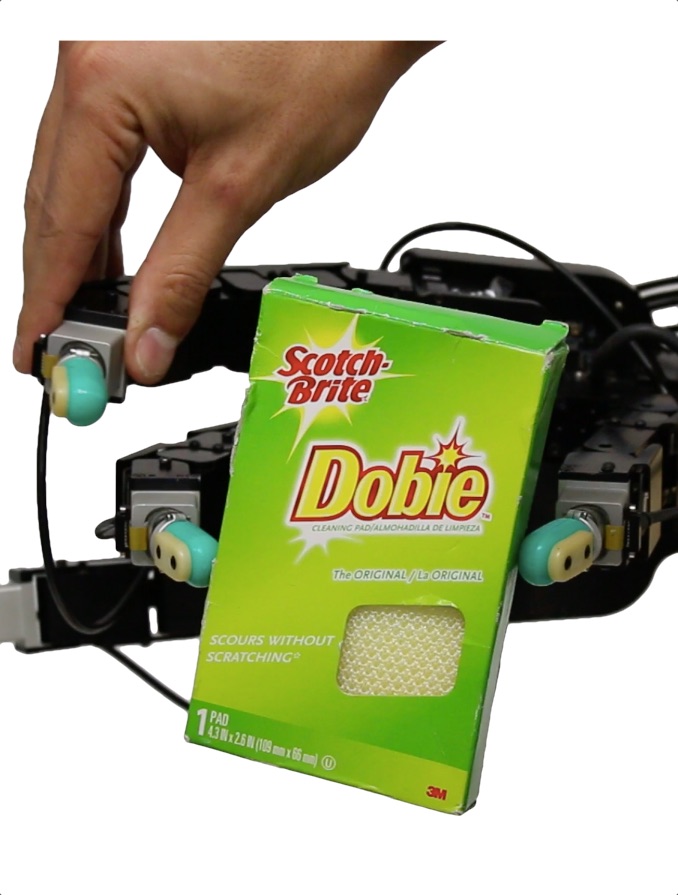}}\hspace{0.001\linewidth}
  \subfloat{\includegraphics[height=0.1\textwidth]{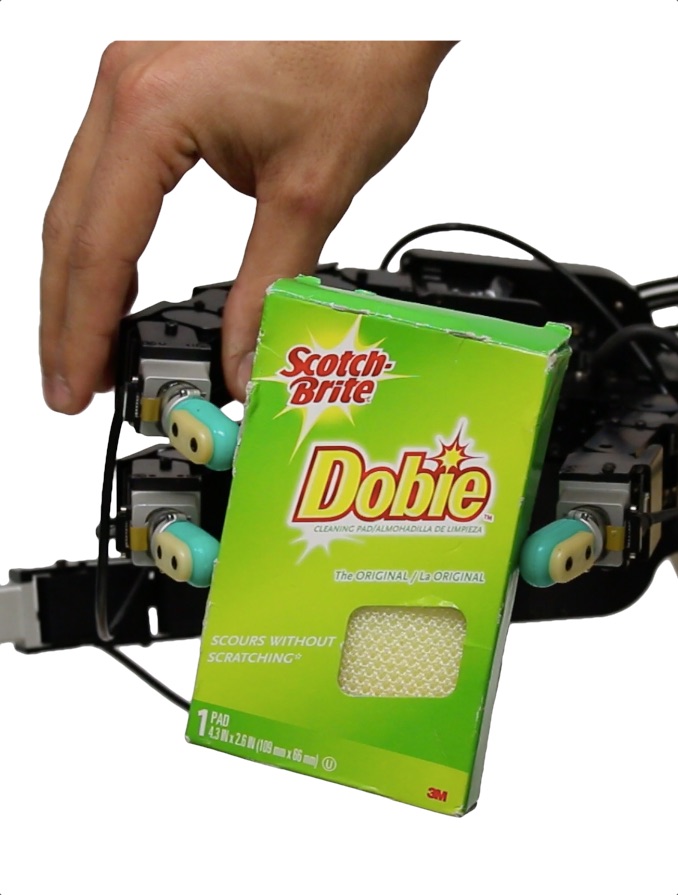}}\hspace{0.001\linewidth}
  \subfloat{\includegraphics[height=0.1\textwidth]{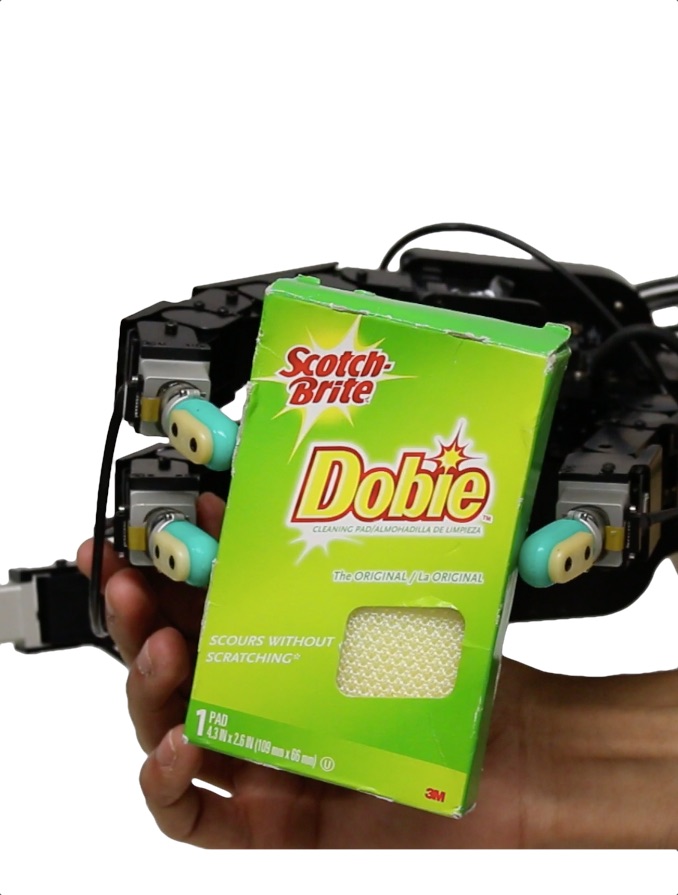}}\hspace{0.001\linewidth}
  \subfloat{\includegraphics[height=0.1\textwidth]{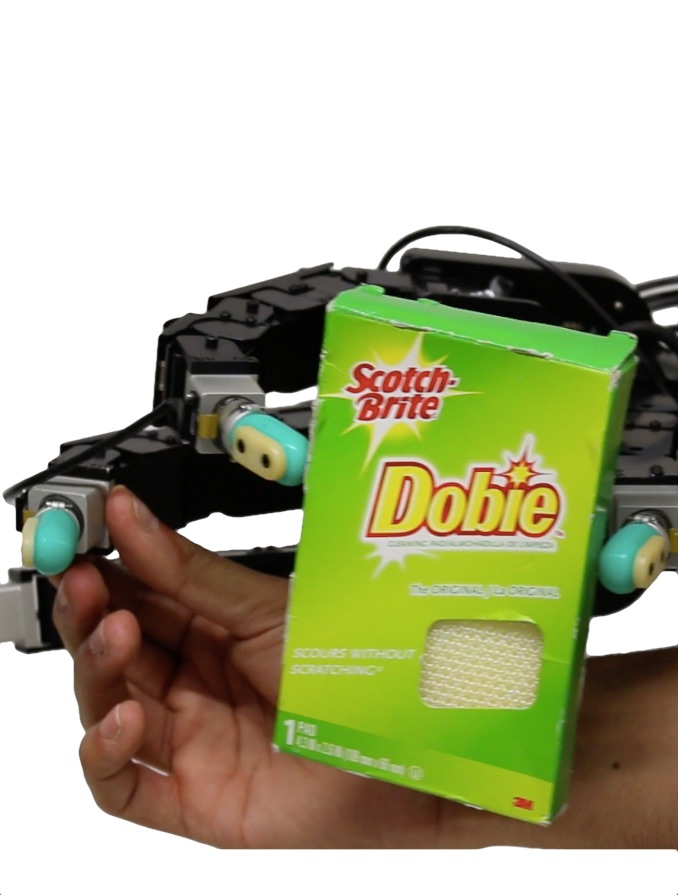}}\hspace{0.001\linewidth}
  \subfloat{\includegraphics[height=0.1\textwidth]{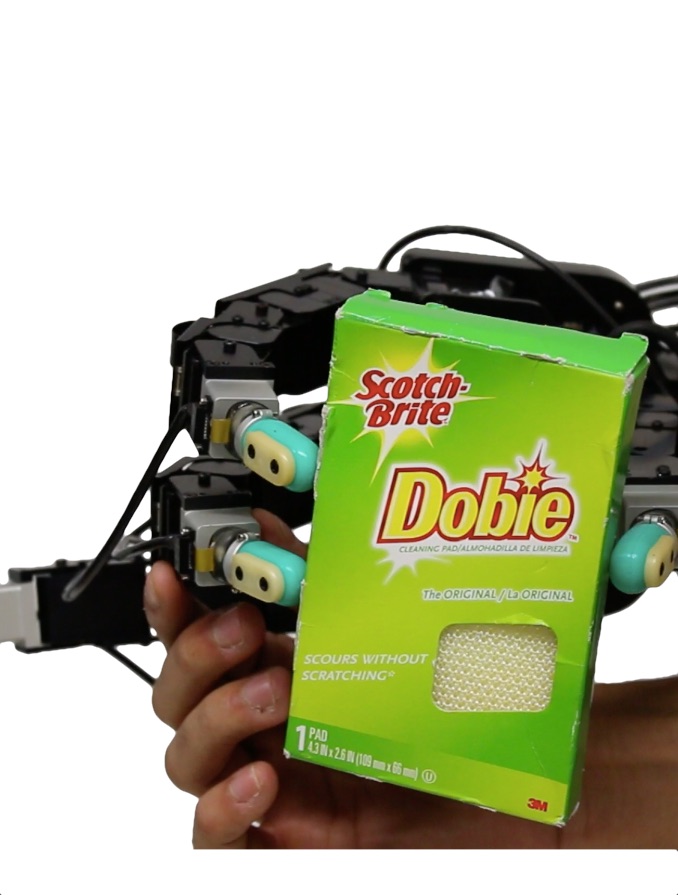}}\hspace{0.001\linewidth}
  \subfloat{\includegraphics[height=0.1\textwidth]{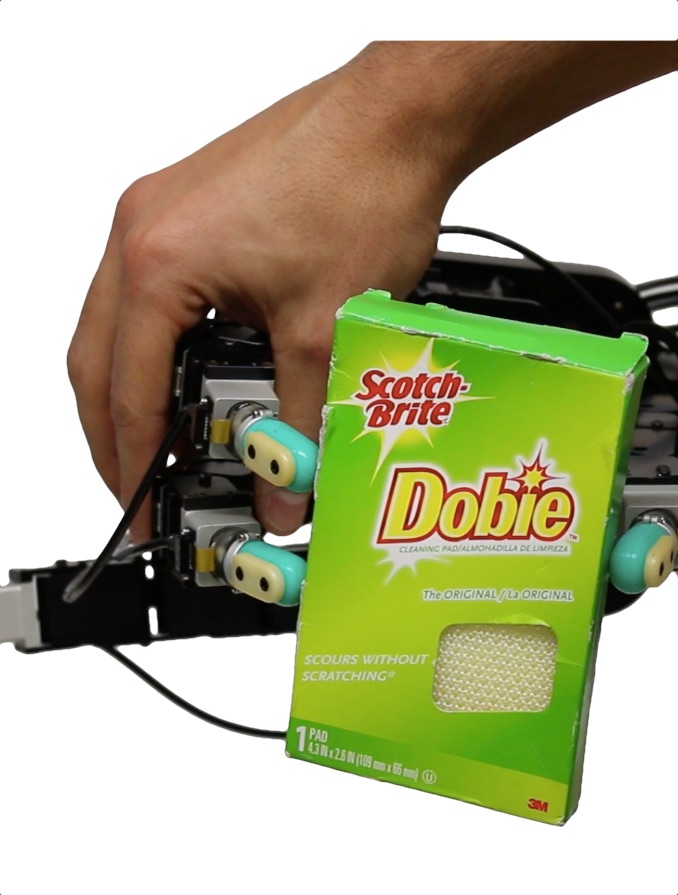}}\hspace{0.001\linewidth}
  \subfloat{\includegraphics[height=0.1\textwidth]{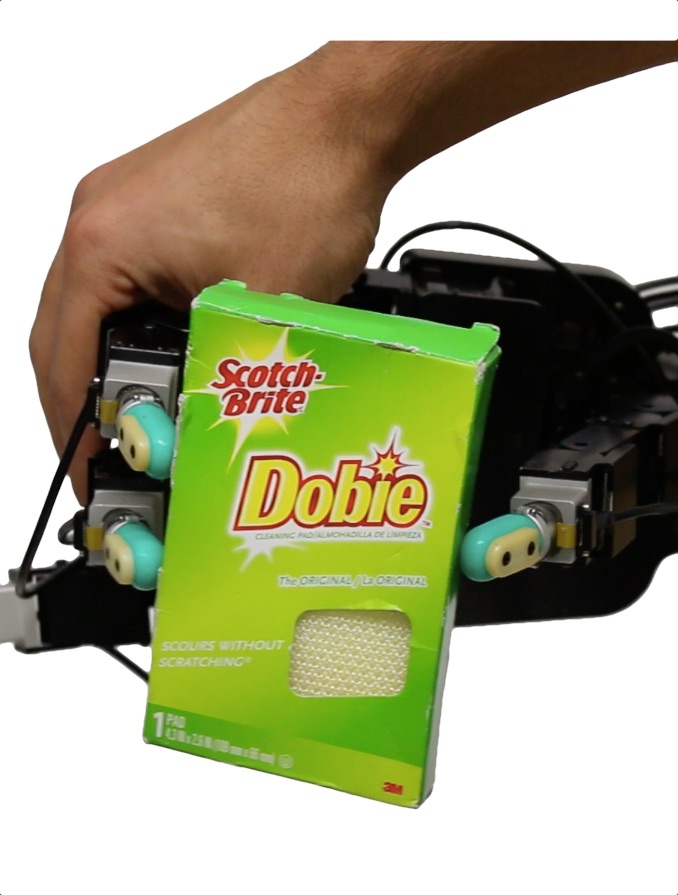}}
\caption{\label{fig:master:slave}Experiments showcasing master slave operation where the fingers stabilize the object despite one finger introducing perturbations in order to change the object's positions in-hand. The experiment showcases how the independent finger grip stabilization controllers, paired with upper level control policies, can enable in-hand object manipulation. Instead of an upper level controller, finger perturbations were introduced by a human experimenter for a two finger grasp (upper row) and a three finger grasp (lower row). In addition, in the three finger grasp we show that fingers can be removed from the object while it is re-stabilized by the remaining fingers.}
\end{figure*}

From the perspective of the independent fingertip controllers, there was no conceptual difference between ’external’ perturbations and those caused by the actions of other fingertips. This interaction was further explored in master-slave experiments during which the experimenter manually pushed or pulled a finger to increase or decrease the force it applied while the controllers of the remaining fingers jointly stabilized the grasp. Indeed, three- and two-digit grasps remained stable even when one of the digits was lifted off the surface of a grasped object. In contrast to more traditional solutions for manipulation control, force sharing between the engaged fingers varied substantially from trial-to-trial due to the emerging nature of the independent finger control policy. Such variability is, however, typical in human manipulation~\cite{edin1992independent, burstedt1997coordination, flanagan1999control, burstedt1999control}. While it could be easily removed by additional regularization, it could actually be beneficial in practice as it allows a wider range of potential solutions (e.g., for use in a manipulation planner). The master-slave experiments are shown in Figure~\ref{fig:master:slave}.

In grasping and in-hand manipulation, instability is synonymous with slip~\cite{tremblay-icra1993}. In this study, we have focused on low-level control of grasp stability rather than finger positioning and re-positioning. However, the results suggest that independent fingertip control at the base level of a hierarchical control framework may enable higher level control policies to perform complex manipulations. In a basic scenario, rotating an object, for instance, would simply require that one of the fingers introduce a desired perturbation to the object, while the remaining fingers keep it stable.

%% file: DiscussionConclusion.tex
\section{Conclusion and Discussion}
\label{sec:conclusion}

The proposed independent finger grip stabilization control approach, inspired by neurophysiological findings, was able to stabilize a wide range of objects by taking advantage of the generalization capabilities of the slip feedback signals. The resulting grasps not only kept the objects stable within the hand but were also robust to perturbations. The adaptability of the approach may also enable higher level control policies to manipulate objects in-hand, as demonstrated by the master-slave experiments.

\subsection{Summary of the Contribution}
\label{sec:summary-of-contributions}
We have corroborated the hypothesis that stable grasps emerge from a set of independent finger controllers. Indeed, the synchronization between fingers emerge from the tactile feedback of each finger controller and enable stable gripping despite disturbances caused by poor contact distribution on the fingertip surfaces, introduced by other fingers action on the object, or external disturbances. Each finger thus automatically compensated for changes that jeopardized grasp stability. Moreover, our modular control approach was shown to be generalizable across multiple objects, even objects that were substantially different from the objects in the training set.

\subsection{Recognized Shortcomings}
\label{sec:current-shortcomings}
Using the low dimensional slip signals defined in previous work \cite{veigastabilizing}, enabled the design of the controller used in this paper. As the full tactile state is much richer than the slip signals, we may potentially have discarded relevant information.

Additionally, in the proposed approach we focused on 'low-level' control of grasp stability. As such, the objects tested were provided to the hand in configurations where the stabilization would be possible, requiring neither finger gating nor re-positioning.

The implemented controller is reactive, albeit that upcoming slips are \emph{predicted} by the controller. The  temporal limitations in this respect have not been analyzed. For comparison, it takes human as much as 60-80 ms to initiate force responses to incipient and overt fingertip slips and at least 50-100 ms to generate substantial counteracting forces \cite{JohWes1999,HagColJoh1996}, i.e., these delays are too long for preventing the loss of a stable grasp once overt slippage occurs.

\subsection{Future Work}

Partitioning the hand into a set of independent fingers allows the manipulation problem to be viewed as a distributed problem where each finger solves the task locally and coordination only emerges by interaction through the object. This setting invites simpler control models than when considering a complete model for the full hand. Specifically, we consider it realistic to use data driven approaches that take into account a richer sensor space, as the dimensionality of the problem is distributed across the fingers. Our future work will focus on exploring the high dimensionality of the feedback signals and learning stabilization controllers using reinforcement learning approaches in these high dimensional spaces.

Our results invite further exploration of master-slave paradigms. In a simple scenario, rotating an object would simply require that one of the fingers introduce a desired perturbation to the object, while the remaining fingers keep it stable.

Finally, for complex manipulations, we posit that independently controlling the fingers will be necessary but not sufficient to achieve robust performance. Using the independent control as the base level in a hierarchical control framework is expected to enable higher level control policies to perform these manipulations, effectively creating a robust control hierarchy, where the task complexity is distributed across the several levels of the hierarchy. Building such a hierarchy is thus a potentially interesting future work.

%% file: Acknowledgements.tex
\section*{Acknowledgments}

The research leading to these results has received funding from the European Community Seventh Framework Programme (FP7/20072013) under grant agreement 610967 (TACMAN), the Intel Corporation and the Swedish Research Council (VR 2016-01635).